\documentclass[11pt]{article}

\usepackage[final]{acl}

\usepackage{times}
\usepackage{latexsym}
\usepackage{amssymb}
\usepackage[T1]{fontenc}

\usepackage[utf8]{inputenc}

\usepackage{microtype}

\usepackage{inconsolata}

\usepackage{graphicx}
\usepackage{multirow}
\usepackage{booktabs}
\usepackage{tabularx}
\usepackage{float}
\usepackage{caption}
\usepackage{subcaption}
\usepackage{enumitem}
\title{Toward a Gricean Retreat: \\Probing LLMs for Knowledge Boundaries and Referent Specificity}

\author{Dananjay Srinivas \qquad Saksham Khatwani \qquad Maria Pacheco \\
        University of Colorado, Boulder \\  \texttt{dasr8731@colorado.edu}
        }

\begin{document}
\maketitle

\begin{abstract}

When asked about entities outside their knowledge boundary, LLMs routinely fabricate plausible-sounding details rather than backing off to safer, more general claims. We frame this failure through a Gricean lens: a cooperative speaker who is uncertain about a referent retreats up the specificity hierarchy, trading informativeness for truthfulness. We ask whether LLMs have the ingredients to perform this retreat. Using a T-REx-based benchmark that varies entity familiarity and referent specificity, we probe models to answer two questions: (i) do their activations encode whether a referent falls inside the knowledge boundary, and (ii) do they anticipate the specificity of the referent they are about to generate? We find that the answer to both is yes, but the two signals are not reconciled in generation. Models overwhelmingly prefer specific referents even when the entity is unknown to them, and do so even when offered correct generic alternatives. The substrate for a Gricean retreat is present, but the policy that would act on it is not. We position our findings as a first step toward Gricean alignment, training or steering objectives that couple knowledge-boundary awareness to referent-specificity during generation.
\end{abstract}

\section{Introduction}\label{sec:introduction}

Large language models (LLMs) have been shown to be capable of eliciting a remarkable amount of knowledge about real-world entities from their parameter space~\cite{petroniLanguageModelsKnowledge2019,veseli-etal-2023-evaluating}. However, this is contingent on the entity being well-represented in pretraining data~\cite{deshmukh-etal-2025-entities}.
Due to the long-tailed and ever-changing nature of real-world entities, it is intractable for LLMs to remain abreast of them all~\cite{huangSurveyHallucinationLarge2025}. This has been characterized in research as the \textit{knowledge boundary problem}, which has been shown to lead to  \textit{factual hallucinations}, i.e., the fabrication of plausible-sounding details about entities outside an LLM’s knowledge boundary~\cite{li-etal-2025-knowledge-boundary,huangSurveyHallucinationLarge2025}.

Humans face this same problem constantly, and can resolve it by offering as much information as possible while staying faithful to what they actually know. \citet{grice1975logic} formalizes this through the \textit{Cooperative Principle}, the implicit norm that interlocutors contribute to a conversation as is required. This principle is realized through several conversational maxims; the two that govern this trade-off most directly are \textit{Quantity} (be as informative as required), and \textit{Quality} (only assert what you know). Consider a speaker who encounters the name Allan Peiper without knowing specifics about him. Drawing on whatever context the name appears in, they can fall back on commonsense category knowledge~\cite{brown_how_1958,cruse_specificity_1977} and refer to him as an Australian, a cyclist, or simply a person. We refer to this move, from a specific referent to a more general one, as a \textit{Gricean retreat}.

\begin{figure}[t]
    \centering
    \includegraphics[width=\linewidth]{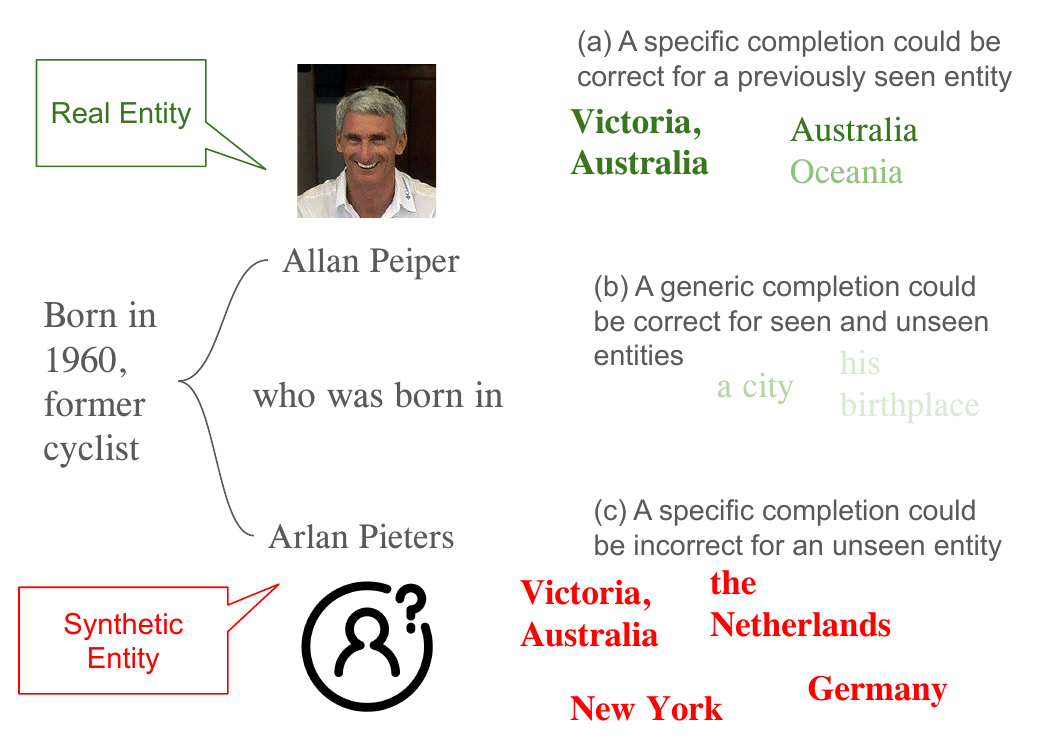}
    \caption{Figure outlines the notion of a \textit{Gricean Retreat}. While dealing with known entities, a specific completion would likely be correct (a), but in the case of unknown entities it would likely lead to an incorrect completion (c). In such cases, models could opt for a less informative, but truthful completion (b).  }
    \label{fig:intro}
\end{figure}

This opens up a natural question for LLMs: when a model is uncertain about a referent, can it perform a Gricean retreat as a cooperative speaker would, moderating the specificity of what it generates to trade informativeness for truthfulness rather than fabricating? Existing approaches to hallucination control largely operate in an \textit{a posteriori} fashion, checking a completed response generation for ``truthfulness'' via internal activation probes~\cite{azaria2023the,marks2024the,li-etal-2025-knowledge-boundary,azizian2025the} and then abstaining or re-generating~\cite{varshney2024a,luo-etal-2024-zero-resource,zhangSirensSongAI2025}. These methods are expensive and all-or-nothing, correcting completed generations rather than calibrating them upfront.


In this paper, we ask whether the ingredients of such upfront calibration are already present inside the model. We curate a benchmark dataset from the T-REx partition~\cite{elsahar-etal-2018-rex} of LAMA~\cite{petroniLanguageModelsKnowledge2019}, covering 8 Wikidata relations across 4 domains (people, corporations, products, and skills). For each relation, we generate three levels of contextual grounding for the subject, synthetic substitutions of the subject to simulate entities outside the model's knowledge boundary, and generic substitutions of the object at varying levels of specificity. We then probe model activations to ask two questions: (i) Does the model internally represent whether a referent is within its knowledge boundary? (ii) Does the model internally represent its upcoming specificity choice, a capability that a Gricean retreat would require?

We find that the answer to both questions is yes, but the two signals are not reconciled in the model's actual generation behavior. Probes can reliably distinguish known from unknown entities, and can predict the specificity of the referent the model is about to generate. Yet when generating, models strongly prefer specific referents regardless of whether the entity falls inside their knowledge boundary, as measured by both perplexity and an extrinsic surprisal elicitation test. In other words, the information needed for a Gricean retreat is present, but the policy that would act on it is not.

We argue this gap represents an untapped opportunity. Rather than treating hallucination as a problem to be caught and corrected after the fact, these internal representations could be leveraged to steer generation toward appropriately general referents when specificity isn't warranted. We see this work as a first step toward \textit{Gricean alignment}, training or steering objectives that explicitly couple knowledge boundary awareness to referent specificity choice during generation. We make the following contributions:


\begin{itemize}[noitemsep, topsep=0pt, leftmargin=*]
\item We construct a benchmark for testing Gricean retreat behavior in LLMs, with varying contextual grounding, synthetic subjects to simulate unknown entities, and generic object substitutions at varying specificity.
\item We show that LLM activations encode both (i) whether an entity falls within the model's knowledge boundary, and (ii) the specificity of the upcoming completion.
\item Despite encoding both signals, models overwhelmingly produce specific referents regardless of boundary status, even under stochastic decoding and when correct generic alternatives are available.
\item We position our findings as a first step toward \textit{Gricean alignment}: objectives that couple knowledge-boundary awareness to specificity choice during generation.
\end{itemize}

\section{Related Work}\label{sec:related-work}

In this section, we outline related findings and situate our contributions in prior scholarship. 

\paragraph{Semantic Hierarchy and Gricean Maxims} 
LLMs have been shown to capture concepts from semantic hierarchy in its representations \cite{hierarchical-concepts,rauba2026deep}
Gricean Maxims underline the implicit assumptions that interlocutors make when conversing with each other -- with two key aspects being ``informativeness'' and ``truthfulness''. \cite{sep-grice} 
We construe the quality of ``informativeness'' as an LLM's preference for hyponymic generation, whereas hypernymic generations will almost always lead to a ``truthful'' response. 
In addition to probing for hierarchical concepts, we tie our findings to a model's ability to control for one quality over another, when contrasted with known and unknown entities. 

\paragraph{Factual Hallucinations due to Knowledge Boundary and Abstention} 
Recognizing LLM knowledge boundary is still a challenging task \cite{huangSurveyHallucinationLarge2025}. Past works have looked at providing context in order to help LLMs deal with entities outside their knowledge \cite{li-etal-2023-large,ren-etal-2025-investigating} or by benchmarking an LLMs ability to know how well they work with entities outside their knowledge boundary \cite{onoe-etal-2022-entity}.
However, these approaches don't setup a policy that could guide LLM behavior when faced with unknown entities without external knowledge or finetuning.
Our approach prescribes LLMs to adhere to the Cooperative Principle when dealing with entities, so as to venture specifics only when a referent entity has been observed in the pretraining data
In a sense, this is similar to abstention \cite{zhangSirensSongAI2025}, but instead of a refusal or hedged response, we favor a $1$-shot truthful generation that is faithful to what the model actually knows. 
Other work that has explored this direction has relied on online-verifying and regenerating answers \cite{varshney2024a,zhao-etal-2024-knowing}, or by testing the understanding of concepts to abstain from generating falsehoods \cite{luo-etal-2024-zero-resource}, both of which are expensive and require multiple passes for a final response.

\paragraph{Probing LLMs to know their knowledge boundary} 
A large body of work has shown extensively that LLMs are capable of representing their truthfulness in their model parameters \cite{marks2024the,azaria2023the,azizian2025the}.
However, to the best of our knowledge, no one has looked into probing LLMs to determine whether it has seen a particular entity in its training data. 
In this work, we use the \texttt{infini-gram API} \cite{Liu2024InfiniGram} to identify distractor entities that an LLM has not come across in its training data, and use them as replacements for known entities. 
To ensure that LLMs are not using shallow heuristics for recall  \cite{saynovaFactRecallHeuristics2025}, we control and test LLMs with distractors that share similar surface properties to the real entity. 
Finally, following \cite{orgad2025llms}, we use linear models  to discriminate between activations obtained from known and unknown entities.

\section{Data}\label{sec:data}


\begin{table}[t]
\resizebox{\columnwidth}{!}{%
\begin{tabular}{clc}
\hline
\textbf{Domain Map} &
  \textbf{Relationships} &
  \textbf{\# Samples} \\ \hline
\begin{tabular}[c]{@{}c@{}}Human\\ $\downarrow$\\ Location\end{tabular} &
  \begin{tabular}[c]{@{}l@{}}P19: [X] was born in [Y]\\ P20: [X] died in [Y]\end{tabular} &
  918 \\ \hline
\begin{tabular}[c]{@{}c@{}}Corporation\\ $\downarrow$\\ Location\end{tabular} &
  \begin{tabular}[c]{@{}l@{}}P740: [X] was founded in [Y]\\ P159: [X] has headquarters in [Y]\end{tabular} &
  808 \\ \hline
\begin{tabular}[c]{@{}c@{}}Product\\ $\downarrow$\\ Corporation\end{tabular} &
  \begin{tabular}[c]{@{}l@{}}P449: [X] was originally aired on [Y]\\ P127: [X] is owned by [Y]\end{tabular} &
  1176 \\ \hline
\begin{tabular}[c]{@{}c@{}}Person\\ $\downarrow$\\ Skill\end{tabular} &
  \begin{tabular}[c]{@{}l@{}}P136: [X] plays [Y] music\\ P413: [X] plays in [Y] position\end{tabular} &
  1590 \\ \hline
\end{tabular}%
}
\caption{The Wikidata properties we utilized in construction of our dataset.}
\label{tab: wiki_relations}
\end{table}

\begin{figure*}[t] 
    
    \hspace{-0.2in}
   \includegraphics[width=1.05\textwidth]{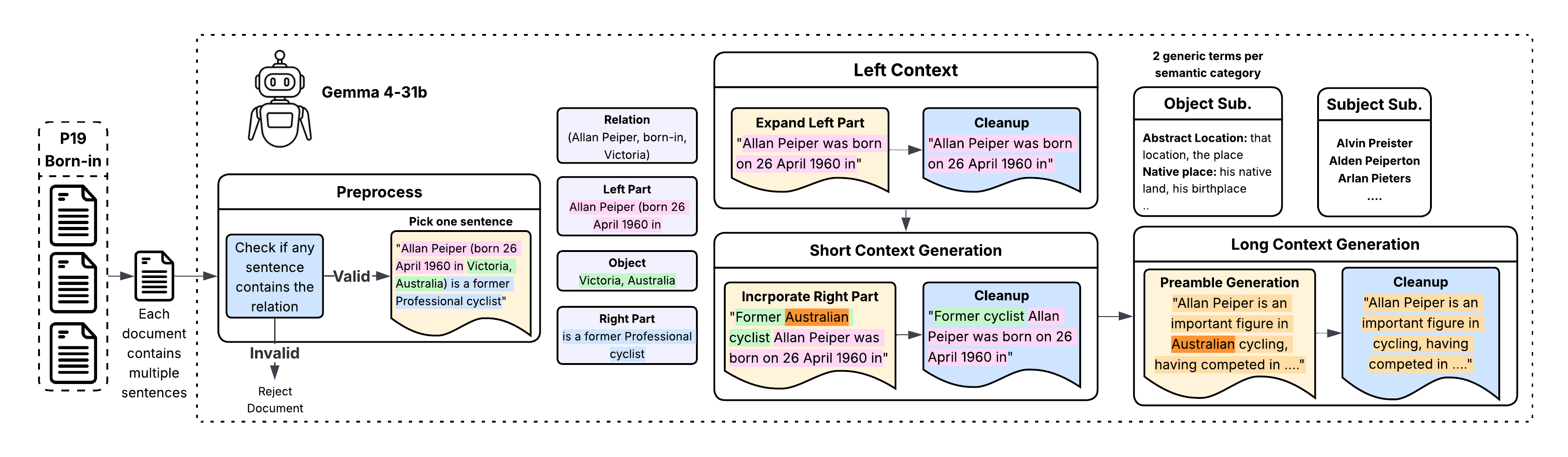}
   \vspace{-.4in}
    \caption{\small This figure depicts the data pipeline stages for P19 relation, which describes the Subject's birthplace. }
    \label{fig:data_pipeline}
\end{figure*}

We construct a benchmark dataset to evaluate the entity–fact elicitation capabilities of LLMs under varying contextual conditions. Given a subject \textbf{\textit{SUB}} associated with an object \textbf{\textit{OBJ}} through a relation \textbf{\textit{REL}}, we generate completion prompts with different amounts of contextual information about \textbf{\textit{SUB}} to assess the model’s ability to recover the corresponding fact. Specifically, we consider three levels of contextualization: (1) minimal context (verbalized relationship), (2) a single-sentence context, and (3) an additional 1–2 sentence preamble describing the subject.

While factual recall can be evaluated using the ground-truth object \textbf{\textit{OBJ}}, assessing the specificity of model generations requires a broader evaluation framework. To this end, we introduce generic substitutions of \textbf{\textit{OBJ}}, enabling analysis of whether the model produces overly generic or semantically diluted responses instead of the precise target fact. Furthermore, to simulate settings in which the subject entity may not be explicitly memorized during pre-training, we generate synthetic substitutions for \textbf{\textit{SUB}}. These components create a diverse evaluation framework for analyzing the Gricean retreat behavior of LLMs across varying levels of contextual grounding and entity familiarity.

We use the T-REx partition \cite{elsahar-etal-2018-rex} of the LAMA dataset \cite{petroniLanguageModelsKnowledge2019} to collect our data.
The source of information in this partition comes from Wikipedia and Wikidata, a repository that is present in common LLM pretraining datasets such as $C4$ \cite{c4} and The Pile \cite{pile}. The Wikipedia source allows us to verify the facts related to the entity being talked about. Of the 46 Wikidata relations in T-REx, we choose a subset of 8 relations that exhibit a many-to-one mapping. This ensures there can only be one right answer when eliciting knowledge about an entity. The full list of relations, along with their sample counts is shown in table~\ref{tab: wiki_relations}.

\subsection{Data Construction Pipeline}
We follow a five-stage pipeline to construct our benchmark from the T-REx dataset. At each stage that requires generation or refinement, we use Gemma 4 31B~\cite{Mesnard2024GemmaOM}. We represent each fact in a relation as a triplet (\textit{\textbf{SUB}}, \textit{\textbf{REL}}, \textit{\textbf{OBJ}}). Below, we describe each stage.

\textit{Stage 1: Preprocess.} For each triplet (\textit{\textbf{SUB}}, \textit{\textbf{REL}}, \textit{\textbf{OBJ}}), T-REx provides multiple candidate sentences. To limit ambiguity, we sample one sentence per triplet that explicitly contains \textit{\textbf{SUB}}, \textit{\textbf{OBJ}}, and the relation \textit{\textbf{REL}} between them, avoiding sentences with only indirect references to \textit{\textbf{REL}}. For example, for the triplet (\textit{Allan Peiper}, \textit{born-in}, \textit{Victoria}), T-REx might contain: (i) "Australian cyclist Allan Peiper has competed in five Tour de France races", and (ii) "Allan Peiper, (born in 26 April, 1960), was a professional cyclist born in Victoria, Australia." We pick the second sentence, as it explicitly mentions the subject's birthplace.


\textit{Stage 2: Context Generation.} In this stage, we generate the three levels of context. For minimal context, we simply express the relationship verbally. An example of minimal context prompt for the previous triplet: "Allan Peiper was born in". For short context generation, we use the complete information in the extracted sentence, utilizing any additional information expressed in the sentence other than the relationship. Both short context and minimal context generation requires reformating the existing information from the sentence. For long context generation, we prompt Gemma to generate a 1-2 sentence preamble in addition to the already generated short context.



\textit{Stage 3: Context Cleanup.} The context generation stage can result in certain contexts which contain direct references about the object. We run the cleanup stage after every context generation stage, where we prompt Gemma with the generated context, and the relation triplet to either replace any direct reference to the object with a generic term or remove the reference all-together. For example, in the short context "Australian cyclist Allan Peiper, having competed in five Tour de France, was born in", the word Australian acts as a direct reference towards the answer "Victoria" and can bias the LLM towards predicting an Australian city.

\textit{Stage 4: Object Substitution.} In this stage, we generate 10 generic substitutions for the object with varying specificity. We provide some in-context examples that we generate using ChatGPT. For example, for "Victoria", and context "Former cyclist Allan Peiper, having competed in five Tour de France, was born in", some of the generic object substitutions could be: "his hometown", "his birthplace", "the country", "a region in his country". 

\textit{Stage 5: Subject Substitution.} The previous stages help us create prompts to test an LLM's knowledge about real subjects. In this stage, we generate synthetic names that replace the subject, simulating the case where the LLM has not encountered the subject entity during pretraining. We prompt Gemma to generate synthetic name, specifically instructing it to generate names with similar backgrounds. For example, if generating substitution for Allan Peiper, we generate similar australian sounding names. As for object substitution, we provide in-context examples for each domain.

To ensure high data generation quality, we refined this pipeline in an incremental fashion. For each relation described in Table~\ref{tab: wiki_relations}, we create specific prompts for every stage stage with relevant in-context examples. We first executed the pipeline for 100 samples, identified the mistakes in every stage and adjusted the corresponding prompts. We continued this cycle till we achieved high quality.

\subsection{Verifying Synthetic Entities}

We verify if our process for artificially creating unseen entities is valid using the \texttt{infini-gram API} \cite{Liu2024InfiniGram}. We use The Pile as our reference corpus because the models we evaluate (Pythia, Section~\ref{sec:intrinsic-analysis}) are trained exclusively on it, which lets us tie entity occurrence in pretraining directly to a model's knowledge boundary.  We do this by sampling 1,000 entities from each relation in our dataset, with an equal number of real and synthetic entities. The \texttt{infini-gram API} provides The Pile dataset in a \textit{train} and \textit{val} split, we check entity occurrences on both of them. 

We observe that real entities are captured in most of the splits, whereas the artificially created entities rarely occur in either split. The median for real entities occurring in the train split across relations range from $112$ - $1989$, however the median ranges for synthetic entities span from $0$-$2$ across all relations. 
The median value of real entities occurring in the val split is between $0$ and $2$ and the median for synthetic entities is always $0$. This distribution of entities for the Corporation-Location relation is presented in Figure~\ref{fig:boxplot_pile_corporation_location}. Figures for all other relations can be found in Appendix~\ref{appendix: DistEntities}.





Ideally, we would like to individually check every real and synthetic entity for its occurrence in the dataset. 
However, there are several artificial entities generated per real entity, which would make it intractable due to the volume of entities. 
Instead, we use this method to validate our process for synthetic entity creation in order to use them for our downstream tasks.

\begin{figure}
    \centering
    \includegraphics[width=\linewidth]{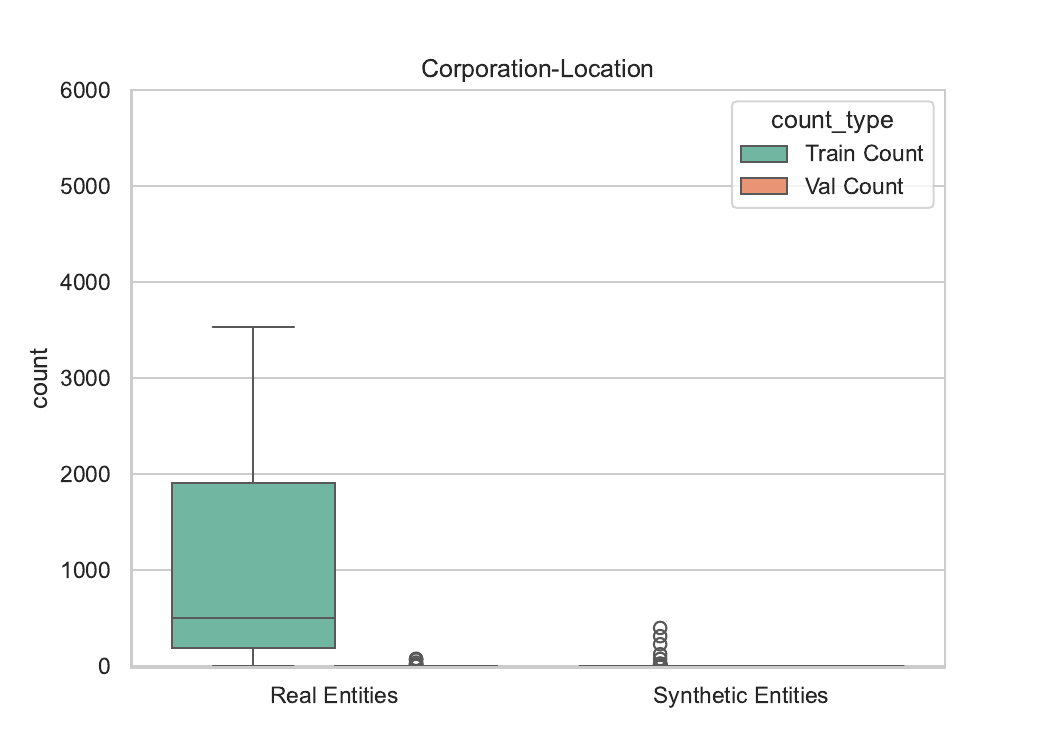}
    \caption{Difference in distribution between the number of real and synthetic entities found in the Pile Dataset for the Corporation-Location domain. The rest of the distributions are provided in the Appendix.}
    \label{fig:boxplot_pile_corporation_location}
\end{figure}




\section{Probing Experiments}\label{sec:intrinsic-analysis}

In this section, we probe models to see if their activations capture whether an entity occurs within the model's knowledge boundary, and whether the model is about to generate a specific or generic completion. We use the Pythia suite~\cite{biderman2023pythiasuiteanalyzinglarge}, testing across the full range of model sizes from 70M to 12B parameters.


\subsection{Method}

\begin{figure}[t]
    \centering
    \includegraphics[width=1\linewidth]{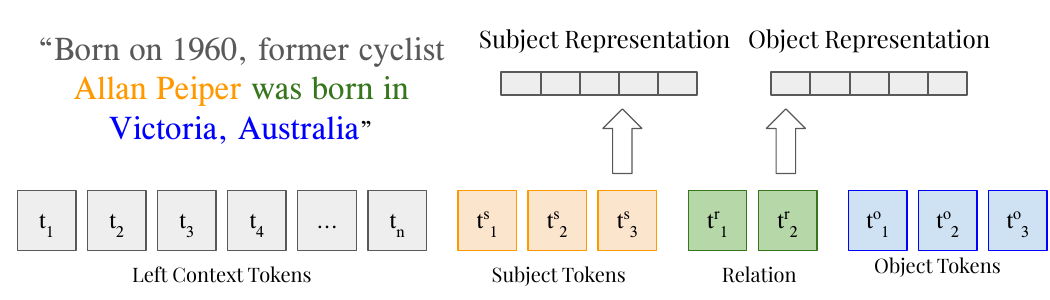}
    \caption{Extracting Subject and Object Representations.}
    \label{fig:repr-extraction}
\end{figure}

Our first probe tests whether the model represents an entity's knowledge boundary status. We extract the hidden activation at the last sub-word token of the entity, which lets the model consider the entire entity before we read out a judgment. We refer to this as the \textit{subject representation}, and use it to predict whether the subject entity is real or synthetic. Our second probe tests whether the model anticipates the specificity of its upcoming completion. We extract the hidden representation at the token position immediately before the completion. We refer to this as the \textit{object representation}, and use it to predict whether the model will produce a specific or generic completion. Figure~\ref{fig:repr-extraction} illustrates the extraction of both representations.


We employ a simple linear probe to inspect the extracted activations following previous work from \cite{marks2024the,azizian2025the}.   
Using $5$-fold Cross Validation, we train a Logistic Regression classifier and calculate the AUROC scores using the Scikit-Learn library in Python \cite{scikit-learn}. 
We then report the average AUROC across all folds.

\paragraph{LLM-as-a-judge to assess completions}

The human annotation cost for assessing LLM completions is high. 
Due to this, we use an LLM-as-a-judge paradigm to label two properties of each completion: its entailment relation to the expected completion, and its specificity level. Entailment captures whether a generation is truthful given the ground truth, allowing for generic responses that are weaker but not wrong. For example, given the ground truth \textit{Victoria}, the completion \textit{Australia} is entailed (Victoria is in Australia) even though it does not lexically match. Specificity captures whether the generation is committal (a specific named entity) or generic (a category-level fall-back). For example, \textit{Victoria} is specific, \textit{a region in Australia} is generic, and \textit{a place} is more generic still. These two labels characterize each completion along the two dimensions a Gricean retreat would calibrate: the specificity of the generation, and its truthfulness given the ground truth. In Section~\ref{sec:extrinsic_analysis} we use these labels to assess whether models actually perform retreats. 

We use \texttt{Deepseek-R1:32b}~\cite{Guo_2025} as the judge, chosen for its reasoning ability and strong performance on NLI benchmarks. To validate this setup, we sample completions from each relation in Section~\ref{sec:data} and have two annotators independently label the entailment and specificity of each completion, allowing us to measure agreement with the LLM-judge. 
The LLM-judge agrees with human annotators at $94.1\%$ for the entailment classification, and $87.4\%$ for the completion specificity annotations. Overall, the LLM agrees with annotators at $90.8\%$. 


Even with the LLM-as-a-judge approach, it is costly to run evaluations across all domains and model sizes. 
We therefore select $2$ models, the \texttt{Pythia-1.4b-deduped} and  \texttt{Pythia-12b-deduped}, to assess completions and identify relations across model sizes. 

\subsection{Findings}

In the interest of brevity, we share the results for just one relation (Person $\rightarrow$ Location). However, the data for the rest of the relations can be found in the appendix (Appendix \ref{appendix: DistEntities}).

\paragraph{Model activations capture entities within and outside its knowledge boundary}

\begin{figure}[t]
    \centering
    \includegraphics[width=\linewidth]{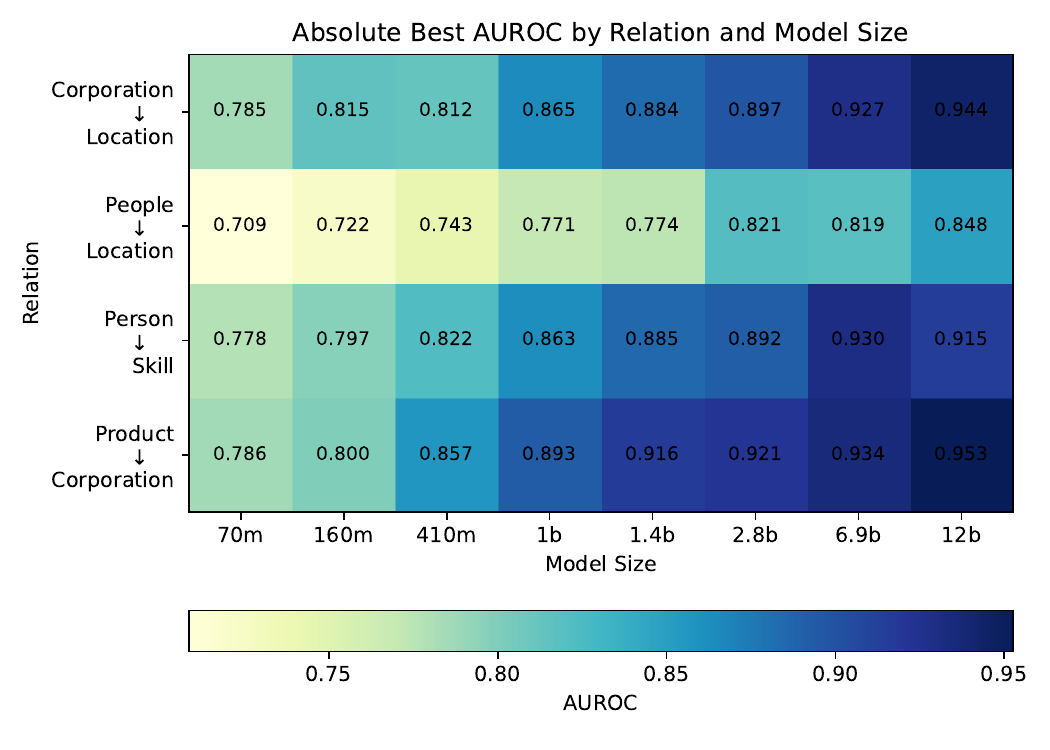}
    \caption{The best AUROC achieved per relation in predicting whether a model has seen an entity, by various model sizes. Each model achieves the highest AUROC in different layers, but they are all roughly just before to the model's middle layer.}
    \label{fig:heatmap-best-auroc}
\end{figure}

\begin{figure}[t]
    \centering
    \includegraphics[width=\linewidth]{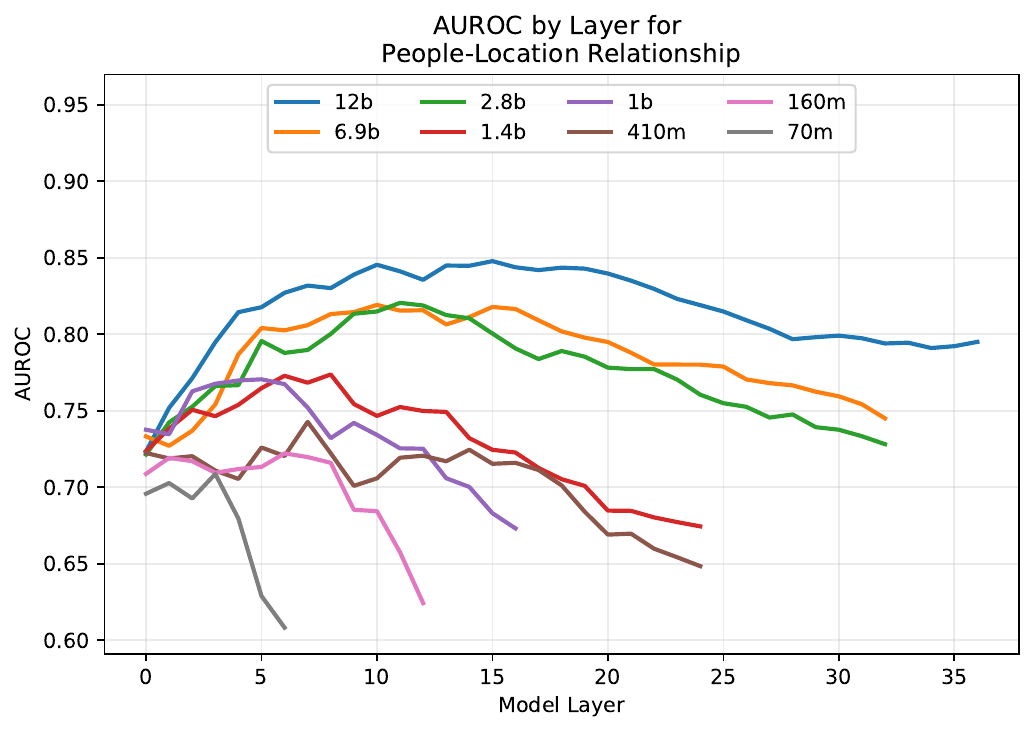}
    \caption{AUROC for predicting whether a model has seen an entity before as it varies with model layers across different models. The difference in layer coverage is due to the fact that smaller models have fewer layers. }
    \label{fig:subj-probe}
\end{figure}

Figure \ref{fig:heatmap-best-auroc} shows that model activations can strongly predict whether an entity has occurred in its pretraining data. 
A linear classifier trained on even the smallest models' activations is able to predict (albeit more weakly) whether an entity is within its knowledge boundary.
Larger models with larger hidden dimensions provide better signal to a linear probe with models with greater than $2$ billion parameters achieving $> 90\%$ AUROC.
Interestingly, the People$\rightarrow$Location relation has the poorest performance, and a potential reason for this is that this relation had the highest number of occurrences of synthetic entities in the Pile dataset. 
This noise may have made it difficult for the linear classifier to predict if an entity occurred in the pretraining data.

Figure \ref{fig:subj-probe} shows that activations from layers immediately before the middle are best at predicting if an entity lies within the models' knowledge boundary.
\citet{azaria2023the} showed that the representation of ``truthfulness'' tends to spike in the middle layers of LLMs, and this may be a parallel phenomenon where a model is able to distinguish a ``truthful'' entity from a false one.

\begin{figure}[]
    \centering
    \includegraphics[width=\linewidth]{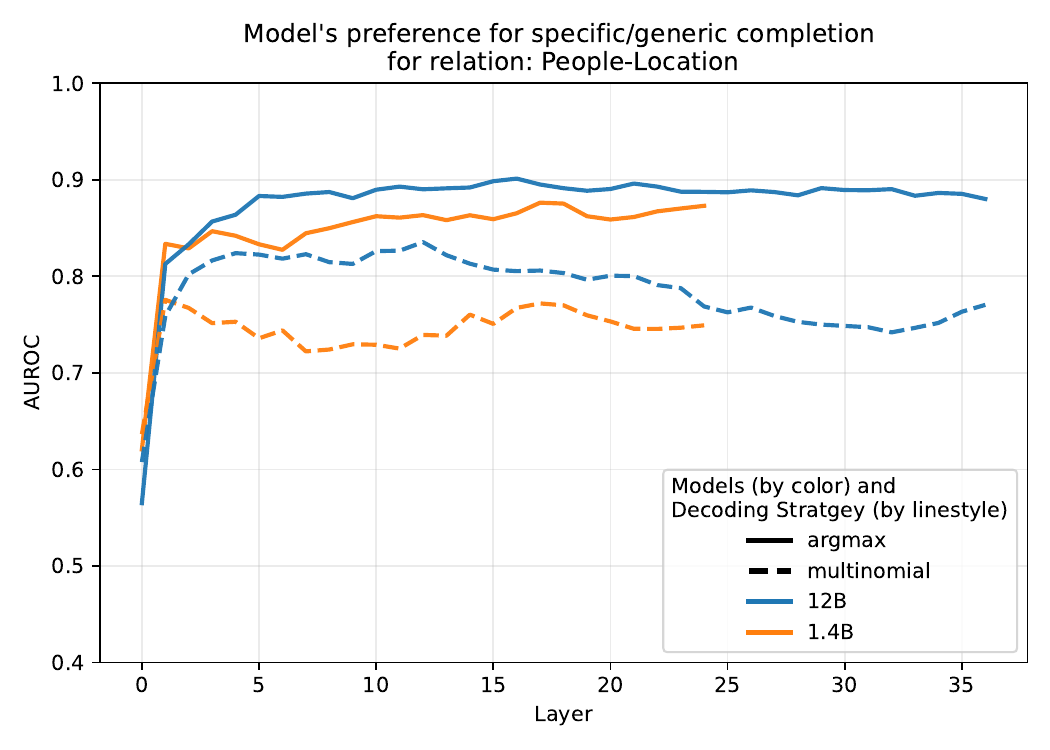}
    \caption{AUROC for predicting whether a model is going to generate a specific or generic completion, as it varies across layers.}
    \label{fig:obj-probe}
\end{figure}

\paragraph{Models activations can reliably predict if the model is going to generate a specific or generic completion} Using the LLM-judge annotations, we sample an equal number of specific and generic completions per relation. The total varies across relations because the model's specificity preference varies across them. Figure~\ref{fig:obj-probe} plots probe AUROC against layer depth. We observe that the probe is unable to predict the model's specificity preference in early layers, achieving AUROC on par with chance. However, predictive accuracy rises strongly as layer depth increases. While both model sizes achieve high AUROC, the 12B model outperforms the 1.4B model in predicting specificity.

We also test two decoding strategies, a greedy argmax, which selects the most probable next token, and multinomial sampling, which samples from the next-token distribution (Figure~\ref{fig:obj-probe}). We observe that activations capture the specificity of argmax decoding more reliably than that of multinomial decoding. This could be due to the deterministic nature of argmax decoding, which may force the model to commit to a specific completion more readily than to a generic one.




\section{Generation Behavior Analysis}\label{sec:extrinsic_analysis}

From our probing experiments, we find that the model encodes signal to identify when an entity is within or outside its knowledge boundary.
We also learned that the model activations can capture whether the model is going to move towards a general or specific completion. 
In this section, we study whether the model uses this information in order to perform \textit{Gricean retreats} by preferring generic completions when faced with entities outside its knowledge boundary.

\subsection{Out-of-the-box Behavior}

We first study how models generate completions without any intervention. The optimal policy in our Gricean framing depends on the entity. For entities within the model's knowledge boundary, the model should produce specific, entailed completions. For entities outside this boundary, the model should retreat to generic but still entailed entities (Figure \ref{fig:intro}). To assess these two dimensions empirically, we use entailment as a proxy for truthfulness and completion specificity as a proxy for informativeness. Our goal is to study whether models are able to balance these two dimensions when generating completions in real and synthetic cases.


We use the LLM-as-a-judge to make this assessment. 
We aggregate results from both model sizes ($12B$ and $1.4B$), and both decoding strategies (argmax and multinomial) for this analysis. 

\begin{figure}[h]
    \centering
    \includegraphics[width=\linewidth]{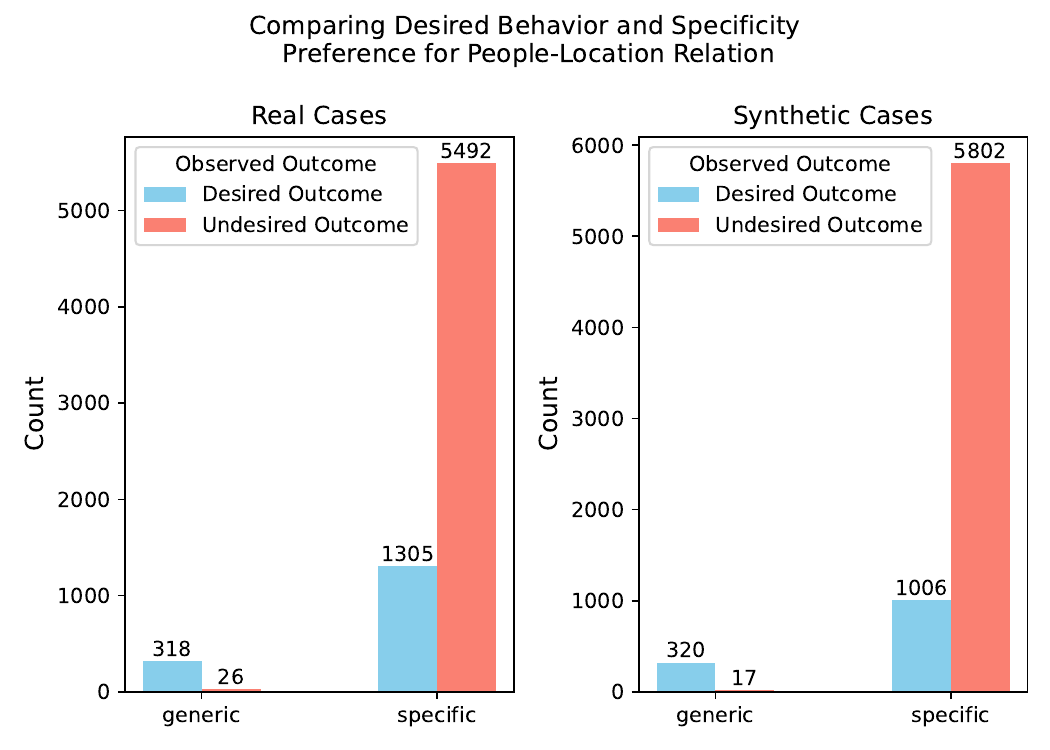}
    \caption{Counts of desired and undesired behaviors across real and synthetic cases. Desired outcomes comprise of truthful completions (though not necessarily informative), and undesired behaviors comprise of falsehoods.}
    \label{fig:nli}
\end{figure}

\paragraph{Results} From Figure \ref{fig:nli}, we observe that in the real and synthetic cases, the model overwhelmingly prefers to be informative at the cost of truthfulness. The model achieves informativeness by generating specific completions.
This helps the model get the right completion frequently when dealing with real cases,
but in the synthetic cases there is no correct specific completion. 
So every time it gives a specific response, it is likely an undesired outcome, unless it generates a specific completion that has a neutral entailment. 

While these observations strongly indicate that the model prefers specific answers in all cases, a potential explanation to this problem could be that the model commits early on to a specific answer and due to nature of a right-to-left decoder, is forced to follow the answer path \cite{azaria2023the,pmlr-v235-zhang24ay}. In order to account for this edge-case, we introduce the next experiment. 

As before, we only disclose one relation studied here. The rest of the plots can be found in the appendix.

\subsection{Surprisal over Candidate Completions}

To test whether the model's preference for specific answers persists when correct generic alternatives are explicitly available, we test if models, when given an option between a correct generic completion and incorrect specific completion, can prefer the correct generic completion. In order to do this, we come up with varying generic-correct answers, and test to see if the average surprisal over the generic statement is lower than a wrong specific completion. The goal here is to see if models can overcome their specific bias if they are presented with correct generic options. 

Since we cannot encode all possible generic answers, we intelligently design generic completions with enough variations that at least some of the synthetic generic answers should be preferred by the model.

\paragraph{Model prefers specific completions, despite having better generic options} 

\begin{figure}
    \centering
    \includegraphics[width=\linewidth]{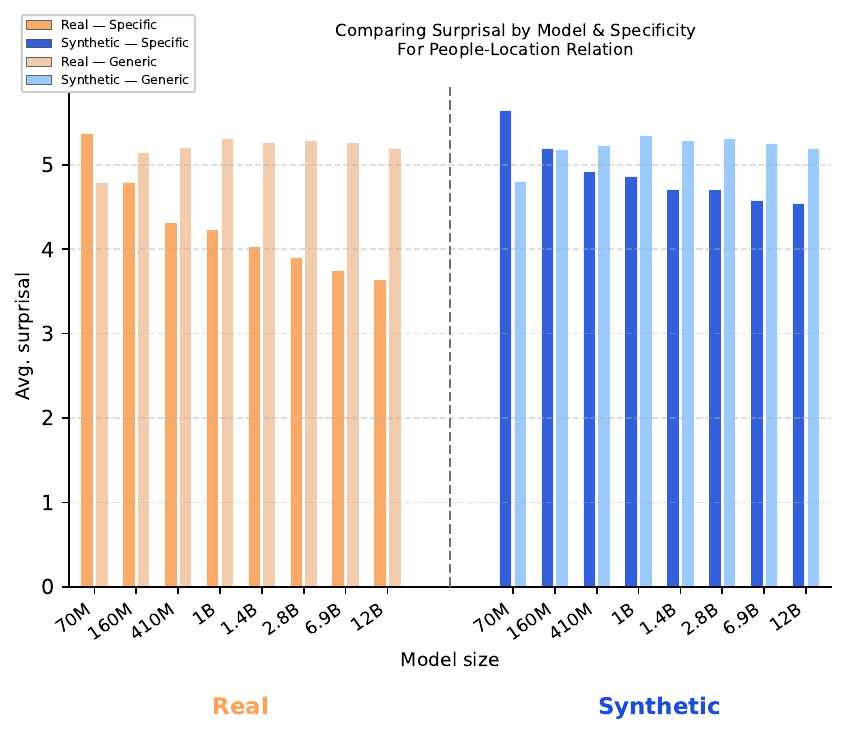}
    \caption{Average surprisal for real and synthetic cases, as it varies across model sizes.}
    \label{fig:extrinsic-analysis}
\end{figure}

The model overwhelming prefers specific completions in real and synthetic scenarios. Figure \ref{fig:extrinsic-analysis} shows average surprisal for real and synthetic cases, across varying model sizes. 
We observe that smaller models actually prefer generic completions, but larger models prefer specific completions.

This could occur because smaller models have smaller hidden dimensions and less layers, so rare specific referents may not be well captured in their parameters, whereas unnamed references constituted by frequent words may be more salient to smaller models.
Larger models may prefer specific completions, as online textual data is usually specific, with the intent of providing a reader with information (e.g., Wikipedia and the News).

\begin{figure}
    \centering
    \includegraphics[width=\linewidth]{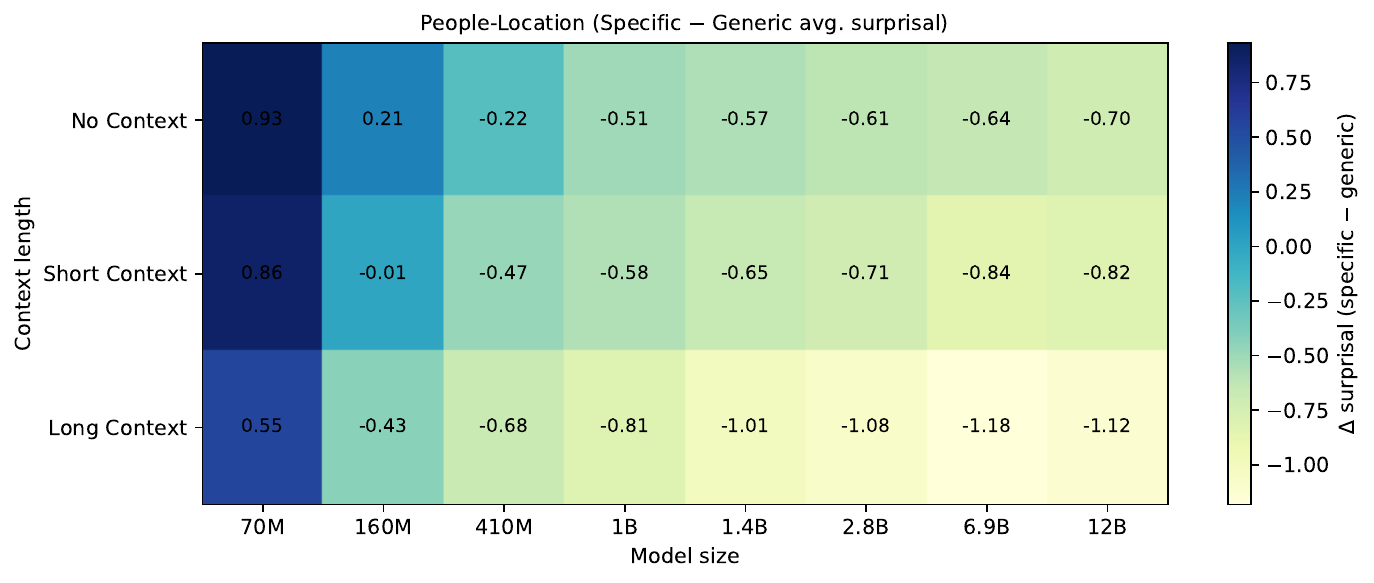}
    \caption{Average difference in specific and generic surprisals.}
    \label{fig:heatmap-surprisal}
\end{figure}

\paragraph{Specificity bias is observed across varying context lengths } We see in Figure \ref{fig:heatmap-surprisal} that specific generations are still preferred across all context lengths, but the preference rises with increasing model size. We also observe that specificity bias increases with increasing context lengths, possibly because the model feels more confident in picking an answer given so much information. 
The long context in synthetic cases  borrows a lot of words from the real cases and could prove to be a distraction. 
This could be a sign of overconfidence on behalf of the model.

\section{Conclusions}\label{sec:conclusions}

In this work, we have outlined a novel strategy for aligning model behaviors to a Gricean paradigm, where the model has to balance its informativeness with its truthfulness. 
In order to do this, we design a data curation process to obtain entities known and unknown to a model by studying its training data. 
Next, we show through linear probing that LLMs' hidden activations can strongly predict whether the model knows if it is real (seen in the training data) or synthetic (unseen to the model). 
We then show that the model knows whether it is going to move towards a specific or a generic generation before the generation. 
Given that the model knows when entities fall within its knowledge boundary, and knowing that it has a strong signal to measure specificity of generation, we test whether this helps the model perform well within our Gricean standard. 
We observe that the model overwhelmingly prefers specific generations, in real and synthetic scenarios. 
We show that this is not just due to a small error at the start of the decoding process, but a problem that persists even when the model is asked to choose between an incorrect specific or a correct generic choice. 
This opens the door for future work in the idea of Gricean alignment of LLMs, to utilize the existing apparatus present in their hidden activations to align them to behave more faithfully to their knowledge.

\section*{Limitations}

Our study is limited by a few resource constraints. 
First, while we checked to ensure that synthetic entities were rarely present in the training data, the effect of the contamination may have skewed our results. 
Second, using an LLM based approach for data generation could further contaminate our data as we are only able to verify for subset of relationships for small set of samples. 
Besides limitations due to data, we were also constrained in the number of samples and models we could test in the LLM-as-a-judge scenario, which means more efforts may be needed to for a more in-depth analysis. 
Lastly, we only consider a subset of relationships and small subset of models, potentially leaving gaps in our analysis. 
We leave addressing these limitations along with the questions regarding implementing Gricean alignment to future work.


\bibliography{custom}

@article{cruse_specificity_1977,
 ISSN = {00222267, 14697742},
 URL = {http://www.jstor.org/stable/4175393},
 author = {D. A. Cruse},
 journal = {Journal of Linguistics},
 number = {2},
 pages = {153--164},
 publisher = {Cambridge University Press},
 title = {The Pragmatics of Lexical Specificity},
 urldate = {2026-05-21},
 volume = {13},
 year = {1977}
}

@article{brown_how_1958,
	Author = {Brown, Roger},
	Journal = {Psychological Review},
	Pages = {14--21},
	Title = {How shall a thing be called?},
	Volume = {65},
	Year = {1958}}

@incollection{grice1975logic,
  address = {New York},
  author = {Grice, H. P.},
  booktitle = {Syntax and Semantics: Vol. 3: Speech Acts},
  pages = {41-58},
  publisher = {Academic Press},
  title = {Logic and Conversation},
  url = {http://www.ucl.ac.uk/ls/studypacks/Grice-Logic.pdf},
  year = 1975
}

@inproceedings{petroniLanguageModelsKnowledge2019,
  title = {Language {{Models}} as {{Knowledge Bases}}?},
  booktitle = {Proceedings of the 2019 {{Conference}} on {{Empirical Methods}} in {{Natural Language Processing}} and the 9th {{International Joint Conference}} on {{Natural Language Processing}} ({{EMNLP-IJCNLP}})},
  author = {Petroni, Fabio and Rockt{\"a}schel, Tim and Riedel, Sebastian and Lewis, Patrick and Bakhtin, Anton and Wu, Yuxiang and Miller, Alexander},
  year = 2019,
  pages = {2463--2473},
  publisher = {Association for Computational Linguistics},
  address = {Hong Kong, China},
  doi = {10.18653/v1/D19-1250},
  urldate = {2025-11-08},
  langid = {english}
}

@inproceedings{veseli-etal-2023-evaluating,
    title = "Evaluating the Knowledge Base Completion Potential of {GPT}",
    author = "Veseli, Blerta  and
      Razniewski, Simon  and
      Kalo, Jan-Christoph  and
      Weikum, Gerhard",
    editor = "Bouamor, Houda  and
      Pino, Juan  and
      Bali, Kalika",
    booktitle = "Findings of the Association for Computational Linguistics: EMNLP 2023",
    month = dec,
    year = "2023",
    address = "Singapore",
    publisher = "Association for Computational Linguistics",
    url = "https://aclanthology.org/2023.findings-emnlp.426/",
    doi = "10.18653/v1/2023.findings-emnlp.426",
    pages = "6432--6443"
}

@article{huangSurveyHallucinationLarge2025,
title = {A {{Survey}} on {{Hallucination}} in {{Large Language Models}}: {{Principles}}, {{Taxonomy}}, {{Challenges}}, and {{Open Questions}}},
shorttitle = {A {{Survey}} on {{Hallucination}} in {{Large Language Models}}},
author = {Huang, Lei and Yu, Weijiang and Ma, Weitao and Zhong, Weihong and Feng, Zhangyin and Wang, Haotian and Chen, Qianglong and Peng, Weihua and Feng, Xiaocheng and Qin, Bing and Liu, Ting},
year = 2025,
month = mar,
journal = {ACM Transactions on Information Systems},
volume = {43},
number = {2},
pages = {1--55},
issn = {1046-8188, 1558-2868},
doi = {10.1145/3703155},
urldate = {2025-11-19},
langid = {english}
}

@inproceedings{deshmukh-etal-2025-entities,
    title = "All Entities are Not Created Equal: Examining the Long Tail for Ultra-Fine Entity Typing",
    author = "Deshmukh, Advait  and
      Umadi, Ashwin  and
      Srinivas, Dananjay  and
      Pacheco, Maria Leonor",
    editor = "Frermann, Lea  and
      Stevenson, Mark",
    booktitle = "Proceedings of the 14th Joint Conference on Lexical and Computational Semantics (*SEM 2025)",
    month = nov,
    year = "2025",
    address = "Suzhou, China",
    publisher = "Association for Computational Linguistics",
    url = "https://aclanthology.org/2025.starsem-1.15/",
    doi = "10.18653/v1/2025.starsem-1.15",
    pages = "189--201",
    ISBN = "979-8-89176-340-1"
}

@inproceedings{li-etal-2025-knowledge-boundary,
    title = "Knowledge Boundary of Large Language Models: A Survey",
    author = "Li, Moxin  and
      Zhao, Yong  and
      Zhang, Wenxuan  and
      Li, Shuaiyi  and
      Xie, Wenya  and
      Ng, See-Kiong  and
      Chua, Tat-Seng  and
      Deng, Yang",
    editor = "Che, Wanxiang  and
      Nabende, Joyce  and
      Shutova, Ekaterina  and
      Pilehvar, Mohammad Taher",
    booktitle = "Proceedings of the 63rd Annual Meeting of the Association for Computational Linguistics (Volume 1: Long Papers)",
    month = jul,
    year = "2025",
    address = "Vienna, Austria",
    publisher = "Association for Computational Linguistics",
    url = "https://aclanthology.org/2025.acl-long.256/",
    doi = "10.18653/v1/2025.acl-long.256",
    pages = "5131--5157",
    ISBN = "979-8-89176-251-0"
}

@inproceedings{
azaria2023the,
title={The Internal State of an {LLM} Knows When It's Lying},
author={Amos Azaria and Tom Mitchell},
booktitle={The 2023 Conference on Empirical Methods in Natural Language Processing},
year={2023},
url={https://openreview.net/forum?id=y2V6YgLaW7}
}

@inproceedings{
marks2024the,
title={The Geometry of Truth: Emergent Linear Structure in Large Language Model Representations of True/False Datasets},
author={Samuel Marks and Max Tegmark},
booktitle={First Conference on Language Modeling},
year={2024},
url={https://openreview.net/forum?id=aajyHYjjsk}
}

@inproceedings{
azizian2025the,
title={The Geometries of Truth Are Orthogonal Across Tasks},
author={Wa{\"\i}ss Azizian and Michael Kirchhof and Eugene Ndiaye and Louis B{\'e}thune and Michal Klein and Pierre Ablin and marco cuturi},
booktitle={ICML 2025 Workshop on Reliable and Responsible Foundation Models},
year={2025},
url={https://openreview.net/forum?id=FdfvGu5rM5}
}

@article{zhangSirensSongAI2025,
  title = {{{Siren}}'s {{Song}} in the {{AI Ocean}}: {{A Survey}} on {{Hallucination}} in {{Large Language Models}}},
  shorttitle = {{{Siren}}'s {{Song}} in the {{AI Ocean}}},
  author = {Zhang, Yue and Li, Yafu and Cui, Leyang and Cai, Deng and Liu, Lemao and Fu, Tingchen and Huang, Xinting and Zhao, Enbo and Zhang, Yu and Chen, Yulong and Wang, Longyue and Luu, Anh Tuan and Bi, Wei and Shi, Freda and Shi, Shuming},
  year = 2025,
  month = sep,
  journal = {Computational Linguistics},
  pages = {1--46},
  issn = {0891-2017},
  doi = {10.1162/COLI.a.16},
  urldate = {2025-11-30}
}

@misc{
varshney2024a,
title={A Stitch in Time Saves Nine: Detecting and Mitigating Hallucinations of {LLM}s by Actively Validating Low-Confidence Generation},
author={Neeraj Varshney and Wenlin Yao and Hongming Zhang and Jianshu Chen and Dong Yu},
year={2024},
url={https://openreview.net/forum?id=d3UGSRLbPo}
}

@inproceedings{luo-etal-2024-zero-resource,
    title = "Zero-Resource Hallucination Prevention for Large Language Models",
    author = "Luo, Junyu  and
      Xiao, Cao  and
      Ma, Fenglong",
    editor = "Al-Onaizan, Yaser  and
      Bansal, Mohit  and
      Chen, Yun-Nung",
    booktitle = "Findings of the Association for Computational Linguistics: EMNLP 2024",
    month = nov,
    year = "2024",
    address = "Miami, Florida, USA",
    publisher = "Association for Computational Linguistics",
    url = "https://aclanthology.org/2024.findings-emnlp.204/",
    doi = "10.18653/v1/2024.findings-emnlp.204",
    pages = "3586--3602"
}

@InCollection{sep-grice,
	author       =	{Grandy, Richard E. and Warner, Richard},
	title        =	{{Paul Grice}},
	booktitle    =	{The {Stanford} Encyclopedia of Philosophy},
	editor       =	{Edward N. Zalta and Uri Nodelman},
	howpublished =	{\url{https://plato.stanford.edu/archives/fall2023/entries/grice/}},
	year         =	{2023},
	edition      =	{{F}all 2023},
	publisher    =	{Metaphysics Research Lab, Stanford University}
}

@ARTICLE{hierarchical-concepts,
  author={Sun, Kai and Bai, Yushi and Tu, Shangqing and Li, Juanzi and Hou, Lei},
  journal={IEEE Transactions on Audio, Speech and Language Processing}, 
  title={Probing Fine-Grained Hierarchical Concept Comprehension and Generation in Large Language Models}, 
  year={2025},
  volume={33},
  number={},
  pages={3229-3242},
  doi={10.1109/TASLPRO.2025.3592339}
  }

@inproceedings{
rauba2026deep,
title={Deep Hierarchical Learning with Nested Subspace Networks for Large Language Models},
author={Paulius Rauba and Mihaela van der Schaar},
booktitle={The Fourteenth International Conference on Learning Representations},
year={2026},
url={https://openreview.net/forum?id=ymUOPsbxLi}
}

@inproceedings{li-etal-2023-large,
    title = "Large Language Models with Controllable Working Memory",
    author = "Li, Daliang  and
      Rawat, Ankit Singh  and
      Zaheer, Manzil  and
      Wang, Xin  and
      Lukasik, Michal  and
      Veit, Andreas  and
      Yu, Felix  and
      Kumar, Sanjiv",
    editor = "Rogers, Anna  and
      Boyd-Graber, Jordan  and
      Okazaki, Naoaki",
    booktitle = "Findings of the Association for Computational Linguistics: ACL 2023",
    month = jul,
    year = "2023",
    address = "Toronto, Canada",
    publisher = "Association for Computational Linguistics",
    url = "https://aclanthology.org/2023.findings-acl.112/",
    doi = "10.18653/v1/2023.findings-acl.112",
    pages = "1774--1793"
}

@inproceedings{ren-etal-2025-investigating,
    title = "Investigating the Factual Knowledge Boundary of Large Language Models with Retrieval Augmentation",
    author = "Ren, Ruiyang  and
      Wang, Yuhao  and
      Qu, Yingqi  and
      Zhao, Wayne Xin  and
      Liu, Jing  and
      Wu, Hua  and
      Wen, Ji-Rong  and
      Wang, Haifeng",
    editor = "Rambow, Owen  and
      Wanner, Leo  and
      Apidianaki, Marianna  and
      Al-Khalifa, Hend  and
      Eugenio, Barbara Di  and
      Schockaert, Steven",
    booktitle = "Proceedings of the 31st International Conference on Computational Linguistics",
    month = jan,
    year = "2025",
    address = "Abu Dhabi, UAE",
    publisher = "Association for Computational Linguistics",
    url = "https://aclanthology.org/2025.coling-main.250/",
    pages = "3697--3715"
}

@inproceedings{onoe-etal-2022-entity,
    title = "Entity Cloze By Date: What {LM}s Know About Unseen Entities",
    author = "Onoe, Yasumasa  and
      Zhang, Michael  and
      Choi, Eunsol  and
      Durrett, Greg",
    editor = "Carpuat, Marine  and
      de Marneffe, Marie-Catherine  and
      Meza Ruiz, Ivan Vladimir",
    booktitle = "Findings of the Association for Computational Linguistics: NAACL 2022",
    month = jul,
    year = "2022",
    address = "Seattle, United States",
    publisher = "Association for Computational Linguistics",
    url = "https://aclanthology.org/2022.findings-naacl.52/",
    doi = "10.18653/v1/2022.findings-naacl.52",
    pages = "693--702"
}

@article{Liu2024InfiniGram,
  title={Infini-gram: Scaling Unbounded n-gram Language Models to a Trillion Tokens},
  author={Liu, Jiacheng and Min, Sewon and Zettlemoyer, Luke and Choi, Yejin and Hajishirzi, Hannaneh},
  journal={arXiv preprint arXiv:2401.17377},
  year={2024}
}

@inproceedings{saynovaFactRecallHeuristics2025,
  title = {Fact {{Recall}}, {{Heuristics}} or {{Pure Guesswork}}? {{Precise Interpretations}} of {{Language Models}} for {{Fact Completion}}},
  shorttitle = {Fact {{Recall}}, {{Heuristics}} or {{Pure Guesswork}}?},
  booktitle = {Findings of the {{Association}} for {{Computational Linguistics}}: {{ACL}} 2025},
  author = {Saynova, Denitsa and Hagstr{\"o}m, Lovisa and Johansson, Moa and Johansson, Richard and Kuhlmann, Marco},
  editor = {Che, Wanxiang and Nabende, Joyce and Shutova, Ekaterina and Pilehvar, Mohammad Taher},
  year = 2025,
  month = jul,
  pages = {18322--18349},
  publisher = {Association for Computational Linguistics},
  address = {Vienna, Austria},
  doi = {10.18653/v1/2025.findings-acl.942},
  urldate = {2025-12-19},
  isbn = {979-8-89176-256-5}
}

@inproceedings{zhao-etal-2024-knowing,
    title = "Knowing What {LLM}s {DO} {NOT} Know: A Simple Yet Effective Self-Detection Method",
    author = "Zhao, Yukun  and
      Yan, Lingyong  and
      Sun, Weiwei  and
      Xing, Guoliang  and
      Meng, Chong  and
      Wang, Shuaiqiang  and
      Cheng, Zhicong  and
      Ren, Zhaochun  and
      Yin, Dawei",
    editor = "Duh, Kevin  and
      Gomez, Helena  and
      Bethard, Steven",
    booktitle = "Proceedings of the 2024 Conference of the North American Chapter of the Association for Computational Linguistics: Human Language Technologies (Volume 1: Long Papers)",
    month = jun,
    year = "2024",
    address = "Mexico City, Mexico",
    publisher = "Association for Computational Linguistics",
    url = "https://aclanthology.org/2024.naacl-long.390/",
    doi = "10.18653/v1/2024.naacl-long.390",
    pages = "7051--7063"
}

@inproceedings{
orgad2025llms,
title={{LLM}s Know More Than They Show: On the Intrinsic Representation of {LLM} Hallucinations},
author={Hadas Orgad and Michael Toker and Zorik Gekhman and Roi Reichart and Idan Szpektor and Hadas Kotek and Yonatan Belinkov},
booktitle={The Thirteenth International Conference on Learning Representations},
year={2025},
url={https://openreview.net/forum?id=KRnsX5Em3W}
}

@inproceedings{elsahar-etal-2018-rex,
    title = "{T}-{RE}x: A Large Scale Alignment of Natural Language with Knowledge Base Triples",
    author = "Elsahar, Hady  and
      Vougiouklis, Pavlos  and
      Remaci, Arslen  and
      Gravier, Christophe  and
      Hare, Jonathon  and
      Laforest, Frederique  and
      Simperl, Elena",
    editor = "Calzolari, Nicoletta  and
      Choukri, Khalid  and
      Cieri, Christopher  and
      Declerck, Thierry  and
      Goggi, Sara  and
      Hasida, Koiti  and
      Isahara, Hitoshi  and
      Maegaard, Bente  and
      Mariani, Joseph  and
      Mazo, H{\'e}l{\`e}ne  and
      Moreno, Asuncion  and
      Odijk, Jan  and
      Piperidis, Stelios  and
      Tokunaga, Takenobu",
    booktitle = "Proceedings of the Eleventh International Conference on Language Resources and Evaluation ({LREC} 2018)",
    month = may,
    year = "2018",
    address = "Miyazaki, Japan",
    publisher = "European Language Resources Association (ELRA)",
    url = "https://aclanthology.org/L18-1544/"
}

@article{pile,
  title={The {P}ile: An 800GB Dataset of Diverse Text for Language Modeling},
  author={Gao, Leo and Biderman, Stella and Black, Sid and Golding, Laurence and Hoppe, Travis and Foster, Charles and Phang, Jason and He, Horace and Thite, Anish and Nabeshima, Noa and Presser, Shawn and Leahy, Connor},
  journal={arXiv preprint arXiv:2101.00027},
  year={2020}
}

@article{c4,
author = {Raffel, Colin and Shazeer, Noam and Roberts, Adam and Lee, Katherine and Narang, Sharan and Matena, Michael and Zhou, Yanqi and Li, Wei and Liu, Peter J.},
title = {Exploring the limits of transfer learning with a unified text-to-text transformer},
year = {2020},
issue_date = {January 2020},
publisher = {JMLR.org},
volume = {21},
number = {1},
issn = {1532-4435},
journal = {J. Mach. Learn. Res.},
month = jan,
articleno = {140},
numpages = {67}
}

@article{Mesnard2024GemmaOM,
  title={Gemma: Open Models Based on Gemini Research and Technology},
  author={Gemma Team Thomas Mesnard and Cassidy Hardin and Robert Dadashi and Surya Bhupatiraju and Shreya Pathak and L. Sifre and Morgane Rivi{\`e}re and Mihir Kale and J Christopher Love and Pouya Dehghani Tafti and L'eonard Hussenot and Aakanksha Chowdhery and Adam Roberts and Aditya Barua and Alex Botev and Alex Castro-Ros and Ambrose Slone and Am'elie H'eliou and Andrea Tacchetti and Anna Bulanova and Antonia Paterson and Beth Tsai and Bobak Shahriari and Charline Le Lan and Christopher A. Choquette-Choo and Cl{\'e}-ment Crepy and Daniel Cer and Daphne Ippolito and David Reid and Elena Buchatskaya and Eric Ni and Eric Noland and Geng Yan and George Tucker and George-Christian Muraru and Grigory Rozhdestvenskiy and Henryk Michalewski and Ian Tenney and Ivan Grishchenko and Jacob Austin and James Keeling and Jane Labanowski and Jean-Baptiste Lespiau and Jeff Stanway and Jenny Brennan and Jeremy Chen and Johan Ferret and Justin Chiu and Justin Mao-Jones and Kather-ine Lee and Kathy Yu and Katie Millican and Lars Lowe Sjoesund and Lisa Lee and Lucas Dixon and Machel Reid and Maciej Mikuła and Mateo Wirth and Michael Sharman and Nikolai Chinaev and Nithum Thain and Olivier Bachem and Os-car Chang and Oscar Wahltinez and Paige Bailey and Paul Michel and Petko Yotov and Pier Giuseppe Sessa and Rahma Chaabouni and Ramona Comanescu and Reena Jana and Rohan Anil and Ross Mcilroy and Ruibo Liu and Ryan Mullins and Samuel L. Smith and Sebastian Borgeaud and Sertan Girgin and Sholto Douglas and Shree Pandya and Siamak Shakeri and Soham De and Ted Klimenko and Tom Hennigan and Vladimir Feinberg and Wojciech Stokowiec and Yu-Hui Chen and Zafarali Ahmed and Zhitao Gong and Tris Warkentin and Ludovic Peran and Minh Giang and Cl{\'e}ment Farabet and Oriol Vinyals and Jeffrey Dean and Koray Kavukcuoglu and Demis Hassabis and Zoubin Ghahramani and Douglas Eck and Joelle Barral and Fernando Pereira and Eli Collins and Armand Joulin and Noah Fiedel and Evan Senter and Alek Andreev and Kathleen Kenealy},
  journal={ArXiv},
  year={2024},
  volume={abs/2403.08295},
  url={https://api.semanticscholar.org/CorpusID:268379206}
}

@misc{biderman2023pythiasuiteanalyzinglarge,
      title={Pythia: A Suite for Analyzing Large Language Models Across Training and Scaling}, 
      author={Stella Biderman and Hailey Schoelkopf and Quentin Anthony and Herbie Bradley and Kyle O'Brien and Eric Hallahan and Mohammad Aflah Khan and Shivanshu Purohit and USVSN Sai Prashanth and Edward Raff and Aviya Skowron and Lintang Sutawika and Oskar van der Wal},
      year={2023},
      eprint={2304.01373},
      archivePrefix={arXiv},
      primaryClass={cs.CL},
      url={https://arxiv.org/abs/2304.01373}, 
}

@article{scikit-learn,
  title={Scikit-learn: Machine Learning in {P}ython},
  author={Pedregosa, F. and Varoquaux, G. and Gramfort, A. and Michel, V.
          and Thirion, B. and Grisel, O. and Blondel, M. and Prettenhofer, P.
          and Weiss, R. and Dubourg, V. and Vanderplas, J. and Passos, A. and
          Cournapeau, D. and Brucher, M. and Perrot, M. and Duchesnay, E.},
  journal={Journal of Machine Learning Research},
  volume={12},
  pages={2825--2830},
  year={2011}
}

@article{Guo_2025,
   title={DeepSeek-R1 incentivizes reasoning in LLMs through reinforcement learning},
   volume={645},
   ISSN={1476-4687},
   url={http://dx.doi.org/10.1038/s41586-025-09422-z},
   DOI={10.1038/s41586-025-09422-z},
   number={8081},
   journal={Nature},
   publisher={Springer Science and Business Media LLC},
   author={Guo, Daya and Yang, Dejian and Zhang, Haowei and Song, Junxiao and Wang, Peiyi and Zhu, Qihao and Xu, Runxin and Zhang, Ruoyu and Ma, Shirong and Bi, Xiao and Zhang, Xiaokang and Yu, Xingkai and Wu, Yu and Wu, Z. F. and Gou, Zhibin and Shao, Zhihong and Li, Zhuoshu and Gao, Ziyi and Liu, Aixin and Xue, Bing and Wang, Bingxuan and Wu, Bochao and Feng, Bei and Lu, Chengda and Zhao, Chenggang and Deng, Chengqi and Ruan, Chong and Dai, Damai and Chen, Deli and Ji, Dongjie and Li, Erhang and Lin, Fangyun and Dai, Fucong and Luo, Fuli and Hao, Guangbo and Chen, Guanting and Li, Guowei and Zhang, H. and Xu, Hanwei and Ding, Honghui and Gao, Huazuo and Qu, Hui and Li, Hui and Guo, Jianzhong and Li, Jiashi and Chen, Jingchang and Yuan, Jingyang and Tu, Jinhao and Qiu, Junjie and Li, Junlong and Cai, J. L. and Ni, Jiaqi and Liang, Jian and Chen, Jin and Dong, Kai and Hu, Kai and You, Kaichao and Gao, Kaige and Guan, Kang and Huang, Kexin and Yu, Kuai and Wang, Lean and Zhang, Lecong and Zhao, Liang and Wang, Litong and Zhang, Liyue and Xu, Lei and Xia, Leyi and Zhang, Mingchuan and Zhang, Minghua and Tang, Minghui and Zhou, Mingxu and Li, Meng and Wang, Miaojun and Li, Mingming and Tian, Ning and Huang, Panpan and Zhang, Peng and Wang, Qiancheng and Chen, Qinyu and Du, Qiushi and Ge, Ruiqi and Zhang, Ruisong and Pan, Ruizhe and Wang, Runji and Chen, R. J. and Jin, R. L. and Chen, Ruyi and Lu, Shanghao and Zhou, Shangyan and Chen, Shanhuang and Ye, Shengfeng and Wang, Shiyu and Yu, Shuiping and Zhou, Shunfeng and Pan, Shuting and Li, S. S. and Zhou, Shuang and Wu, Shaoqing and Yun, Tao and Pei, Tian and Sun, Tianyu and Wang, T. and Zeng, Wangding and Liu, Wen and Liang, Wenfeng and Gao, Wenjun and Yu, Wenqin and Zhang, Wentao and Xiao, W. L. and An, Wei and Liu, Xiaodong and Wang, Xiaohan and Chen, Xiaokang and Nie, Xiaotao and Cheng, Xin and Liu, Xin and Xie, Xin and Liu, Xingchao and Yang, Xinyu and Li, Xinyuan and Su, Xuecheng and Lin, Xuheng and Li, X. Q. and Jin, Xiangyue and Shen, Xiaojin and Chen, Xiaosha and Sun, Xiaowen and Wang, Xiaoxiang and Song, Xinnan and Zhou, Xinyi and Wang, Xianzu and Shan, Xinxia and Li, Y. K. and Wang, Y. Q. and Wei, Y. X. and Zhang, Yang and Xu, Yanhong and Li, Yao and Zhao, Yao and Sun, Yaofeng and Wang, Yaohui and Yu, Yi and Zhang, Yichao and Shi, Yifan and Xiong, Yiliang and He, Ying and Piao, Yishi and Wang, Yisong and Tan, Yixuan and Ma, Yiyang and Liu, Yiyuan and Guo, Yongqiang and Ou, Yuan and Wang, Yuduan and Gong, Yue and Zou, Yuheng and He, Yujia and Xiong, Yunfan and Luo, Yuxiang and You, Yuxiang and Liu, Yuxuan and Zhou, Yuyang and Zhu, Y. X. and Huang, Yanping and Li, Yaohui and Zheng, Yi and Zhu, Yuchen and Ma, Yunxian and Tang, Ying and Zha, Yukun and Yan, Yuting and Ren, Z. Z. and Ren, Zehui and Sha, Zhangli and Fu, Zhe and Xu, Zhean and Xie, Zhenda and Zhang, Zhengyan and Hao, Zhewen and Ma, Zhicheng and Yan, Zhigang and Wu, Zhiyu and Gu, Zihui and Zhu, Zijia and Liu, Zijun and Li, Zilin and Xie, Ziwei and Song, Ziyang and Pan, Zizheng and Huang, Zhen and Xu, Zhipeng and Zhang, Zhongyu and Zhang, Zhen},
   year={2025},
   month=Sept, pages={633–638} }

@InProceedings{pmlr-v235-zhang24ay,
  title = 	 {How Language Model Hallucinations Can Snowball},
  author =       {Zhang, Muru and Press, Ofir and Merrill, William and Liu, Alisa and Smith, Noah A.},
  booktitle = 	 {Proceedings of the 41st International Conference on Machine Learning},
  pages = 	 {59670--59684},
  year = 	 {2024},
  editor = 	 {Salakhutdinov, Ruslan and Kolter, Zico and Heller, Katherine and Weller, Adrian and Oliver, Nuria and Scarlett, Jonathan and Berkenkamp, Felix},
  volume = 	 {235},
  series = 	 {Proceedings of Machine Learning Research},
  month = 	 {21--27 Jul},
  publisher =    {PMLR},
  url = 	 {https://proceedings.mlr.press/v235/zhang24ay.html}
}

\appendix

\section{Additional plots} \label{appendix: DistEntities}
\begin{figure}[h!]
    \centering
    \includegraphics[width=\linewidth]{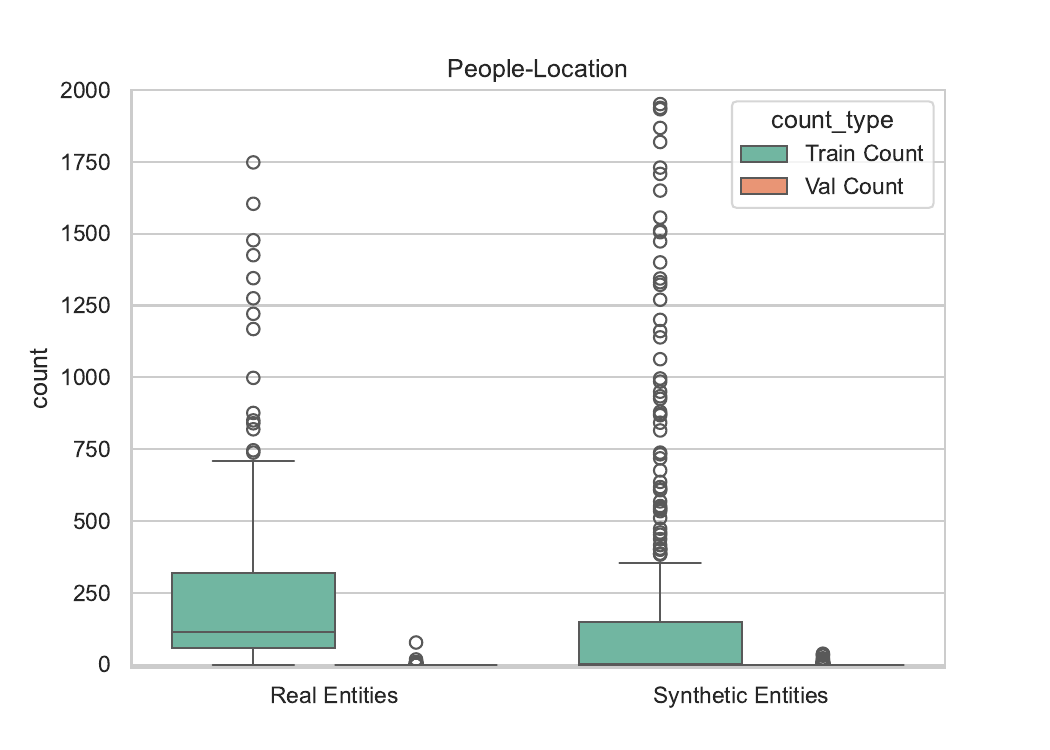}
    \caption{Difference in distribution between the number of real and synthetic entities found in the Pile Dataset for the People-Location domain.}
\end{figure}

\begin{figure}[h!]
    \centering
    \includegraphics[width=\linewidth]{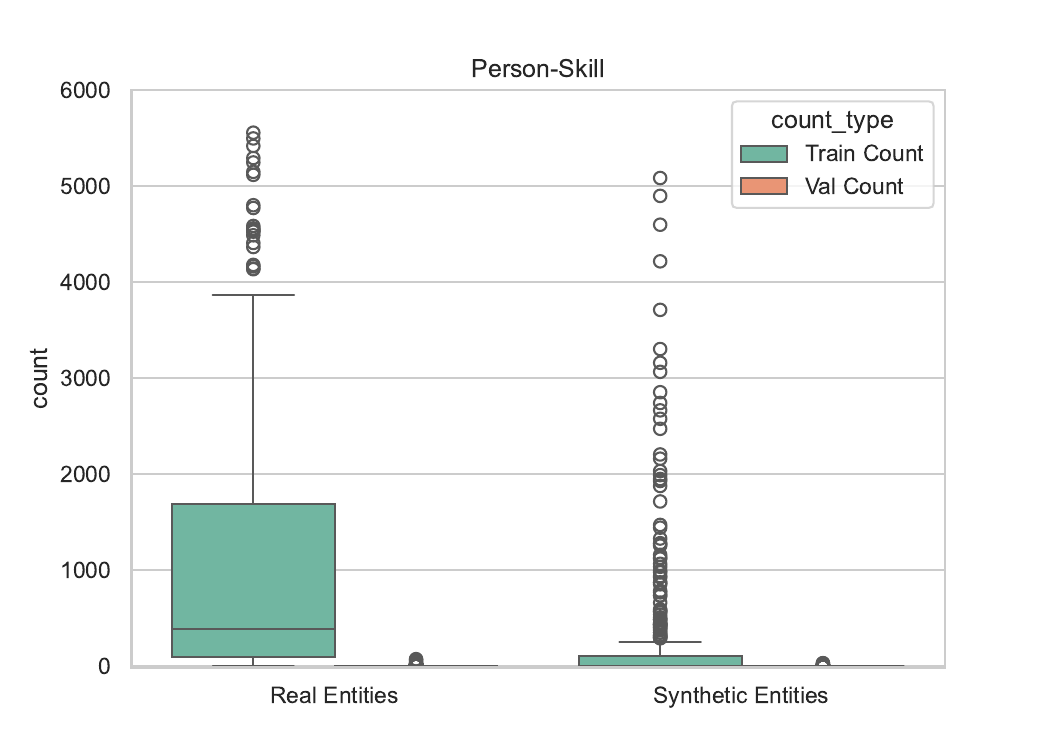}
    \caption{Difference in distribution between the number of real and synthetic entities found in the Pile Dataset for the People-Skill domain.}
\end{figure}

\begin{figure}[h]
    \centering
    \includegraphics[width=\linewidth]{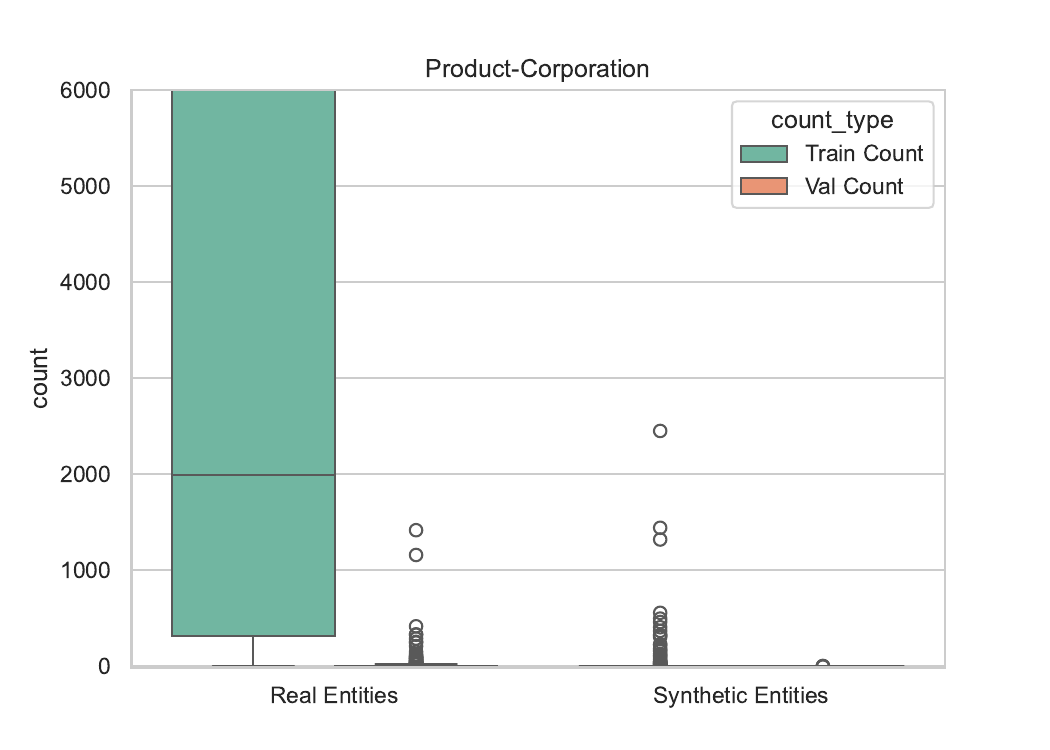}
    \caption{Difference in distribution between the number of real and synthetic entities found in the Pile Dataset for the Product-Corporation domain.}
\end{figure}


\begin{figure}[h!]
    \centering
    \includegraphics[width=\linewidth]{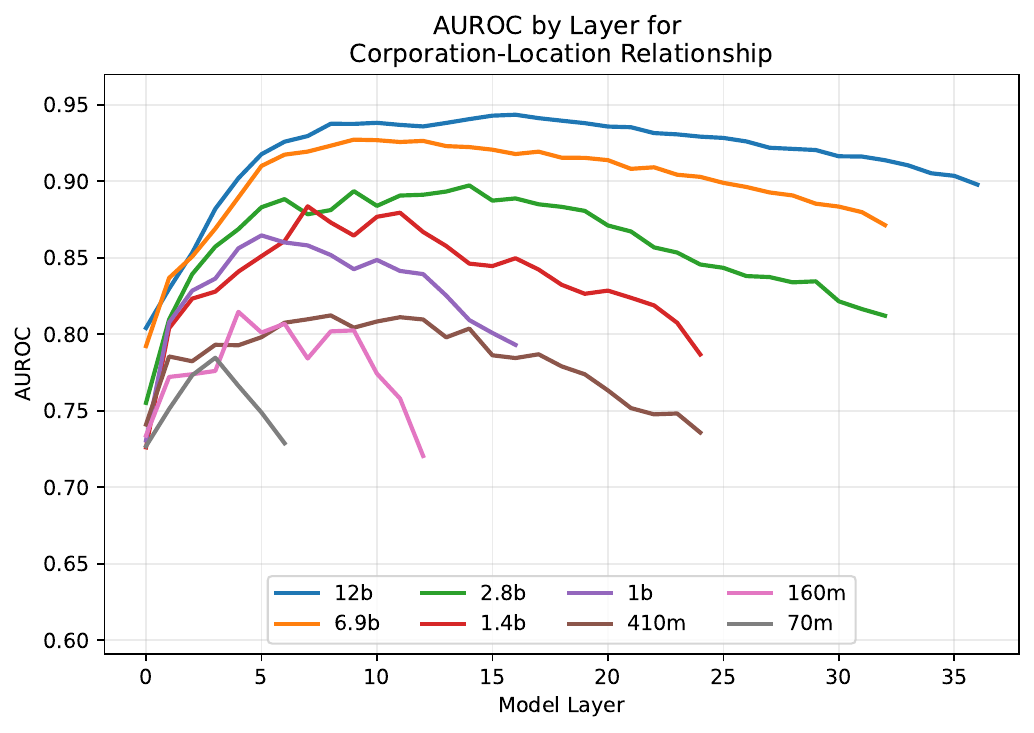}
    \caption{AUROC by layer for Corporation-Location Relationship}
\end{figure}

\begin{figure}[h!]
    \centering
    \includegraphics[width=\linewidth]{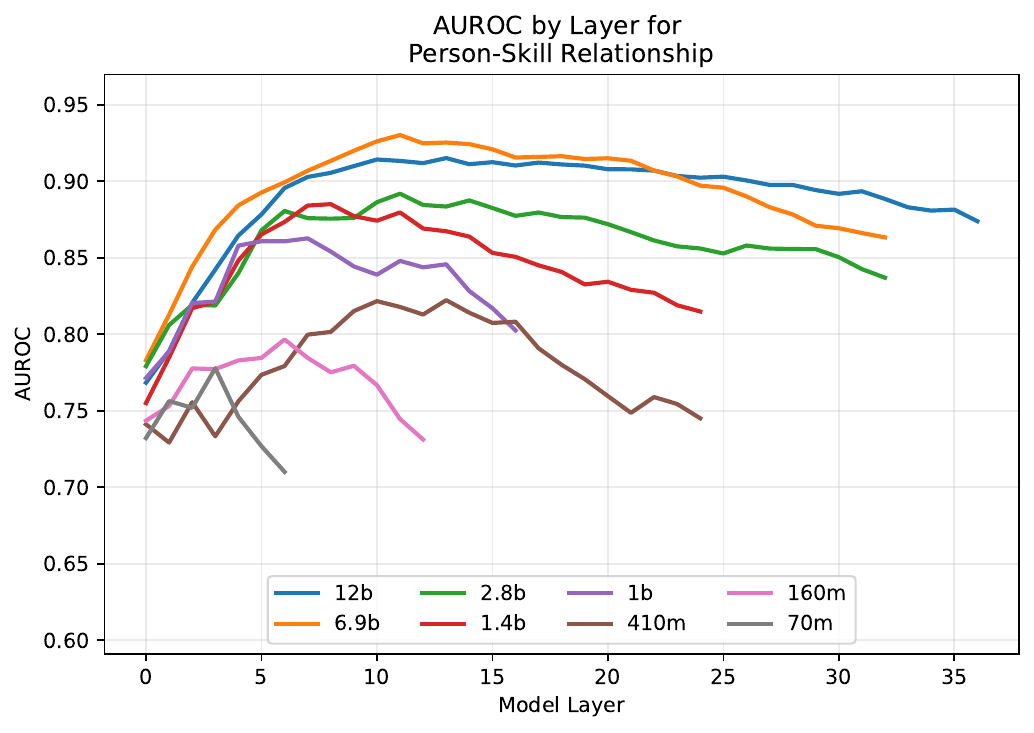}
    \caption{AUROC by layer for People-Skill Relationship}
\end{figure}

\begin{figure}[h!]
    \centering
    \includegraphics[width=\linewidth]{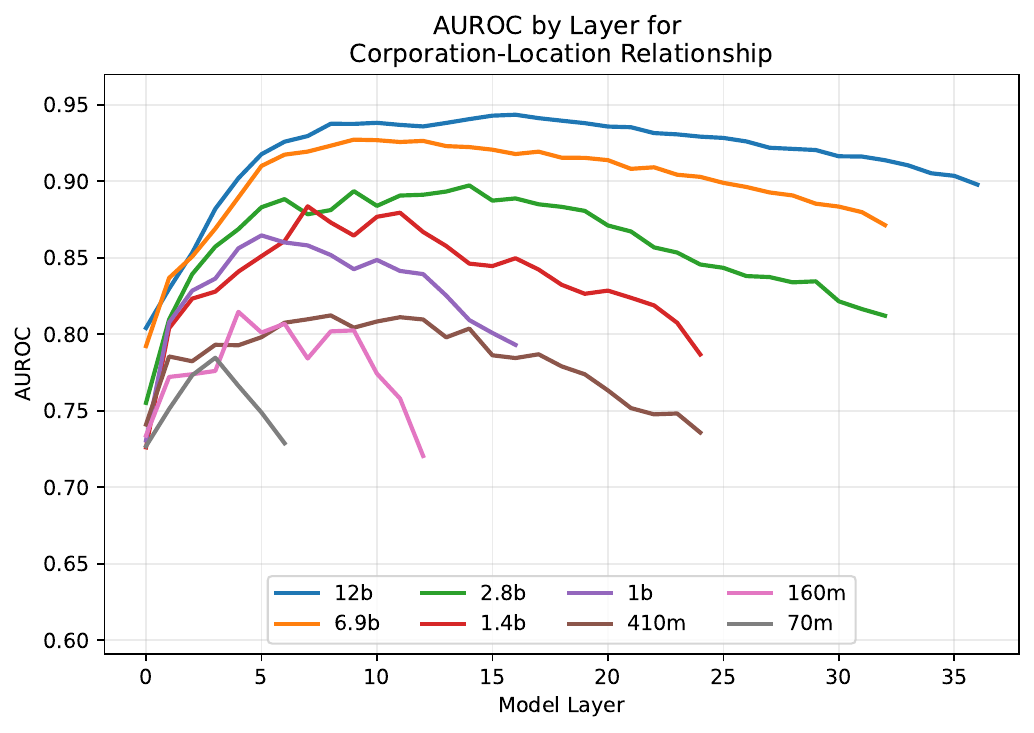}
    \caption{AUROC by layer for Corporation-Location Relationship}
\end{figure}


\begin{figure}[h]
    \centering
    \includegraphics[width=\linewidth]{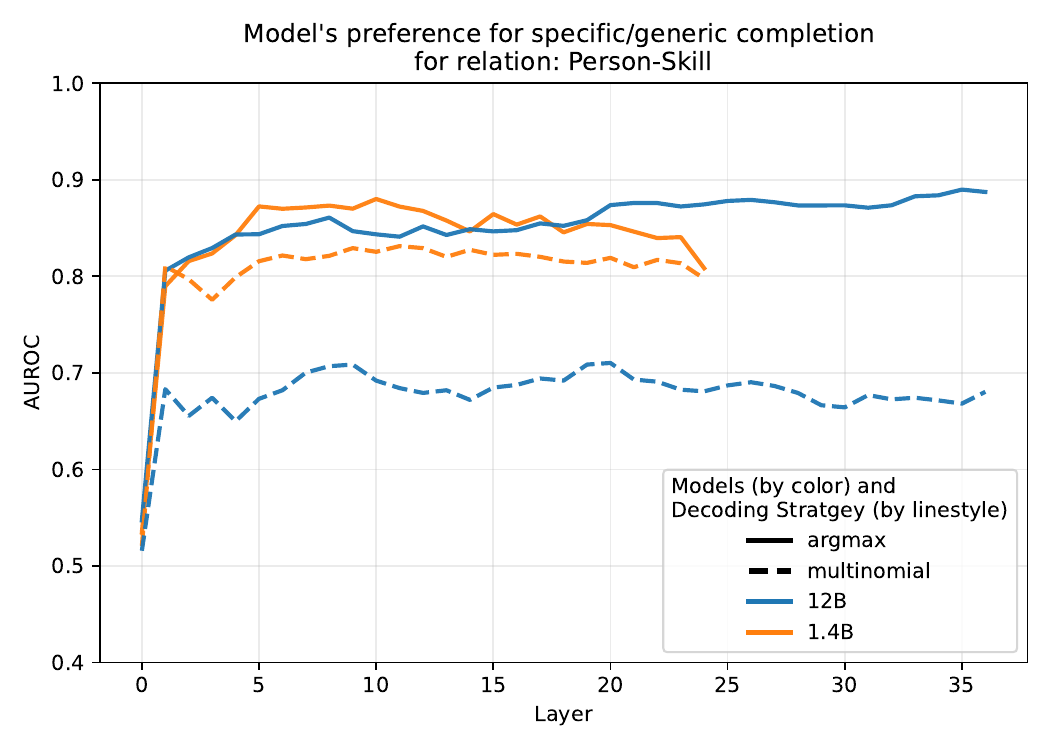}
    \caption{AUROC for predicting specific/generic completion for Person-Skill relation.}
\end{figure}

\begin{figure}[h]
    \centering
    \includegraphics[width=\linewidth]{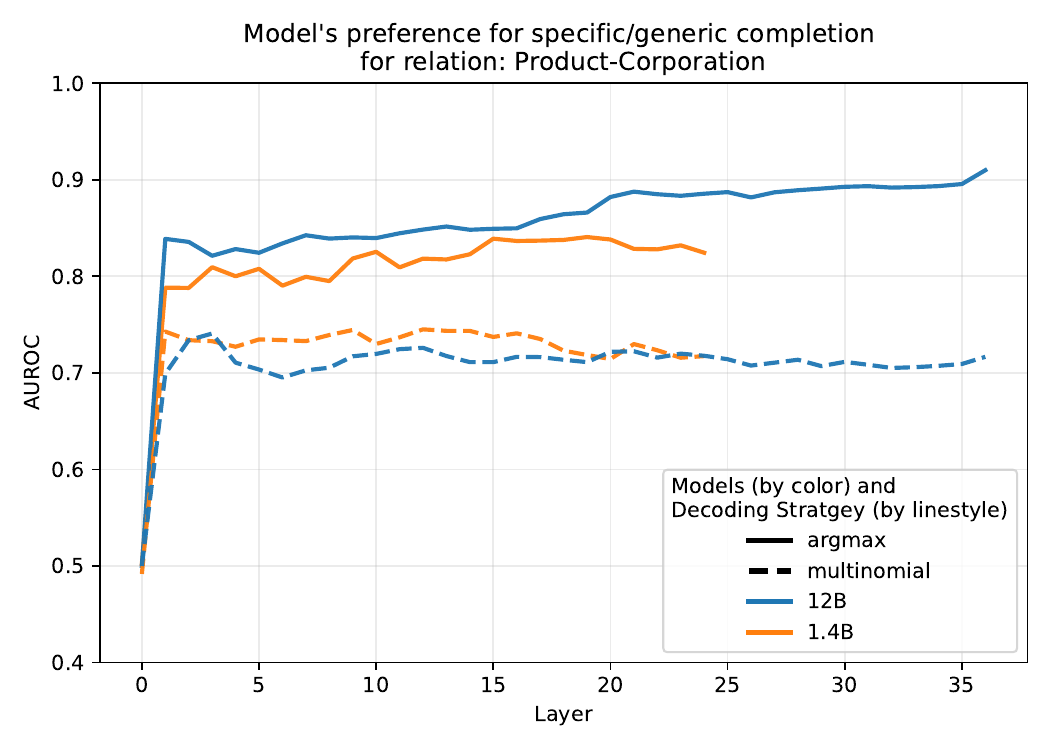}
    \caption{AUROC for predicting specific/generic completion for Production-Corporation relation.}
\end{figure}

\begin{figure}[h]
    \centering
    \includegraphics[width=\linewidth]{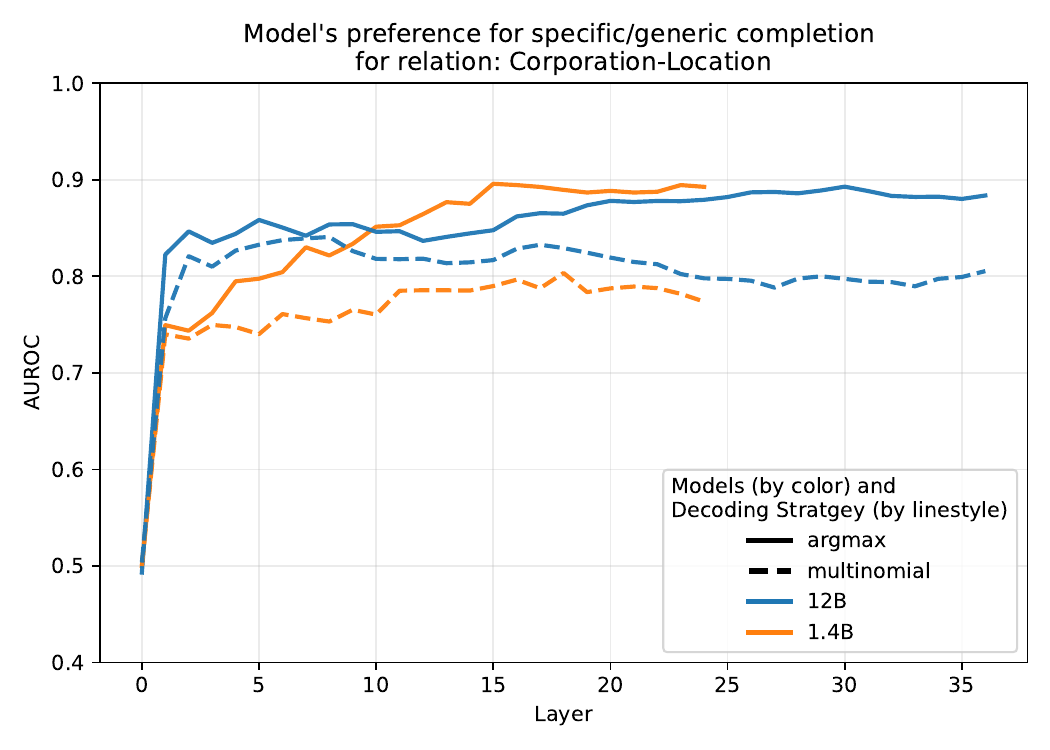}
    \caption{AUROC for predicting specific/generic completion for Corporation-Location relation.}
\end{figure}


\begin{figure}[h]
    \centering
    \includegraphics[width=\linewidth]{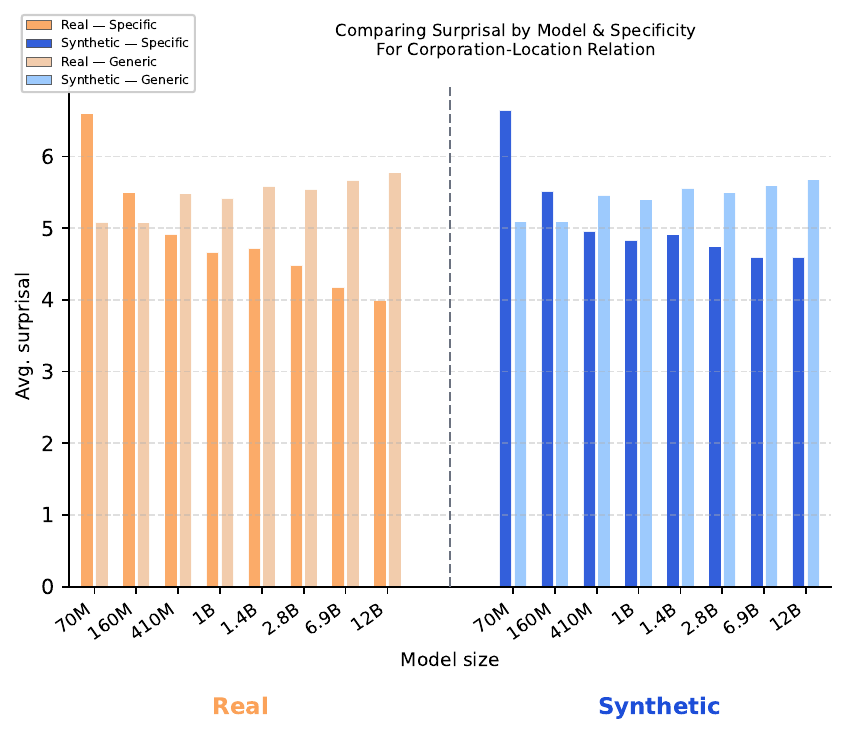}
    \caption{Surprisal comparison per Model and Specificity for Corporation-Location relation.}
\end{figure}

\begin{figure}[h]
    \centering
    \includegraphics[width=\linewidth]{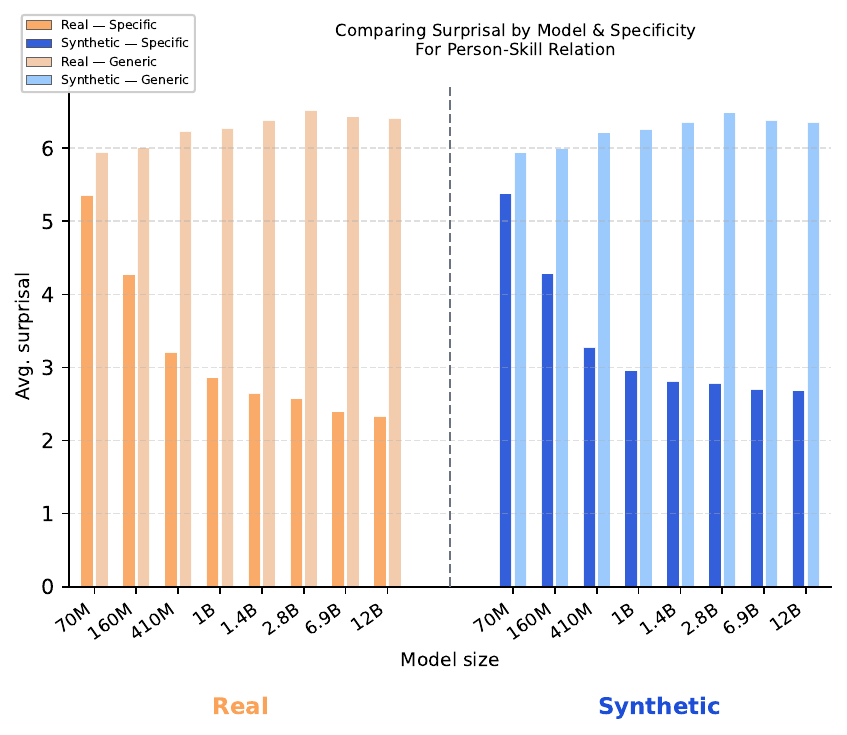}
    \caption{Surprisal comparison per Model and Specificity for People-Skill relation.}
\end{figure}

\begin{figure}[h]
    \centering
    \includegraphics[width=\linewidth]{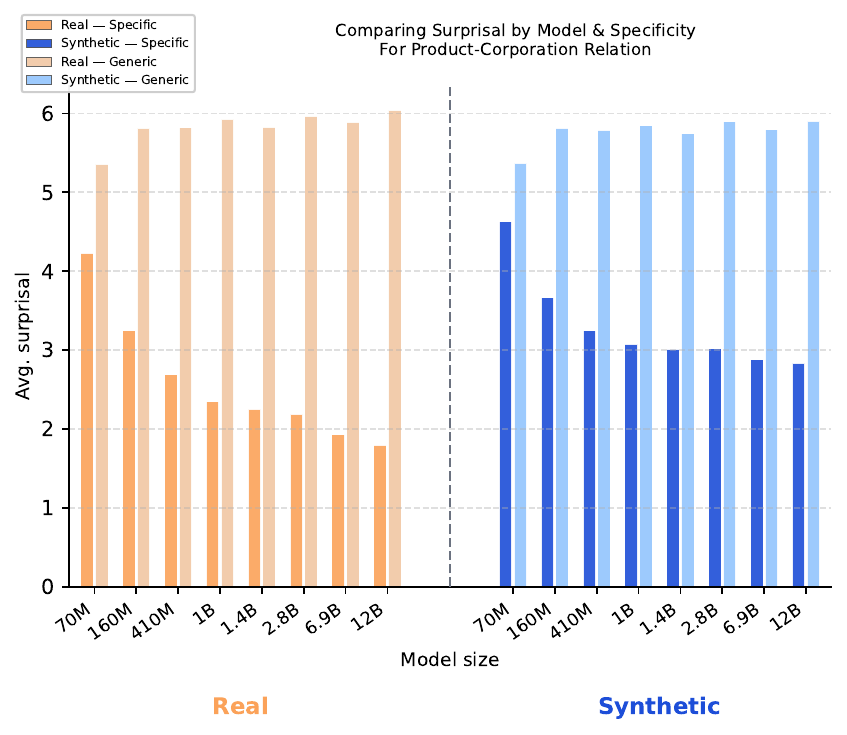}
    \caption{Surprisal comparison per Model and Specificity for Product-Corporation relation.}
\end{figure}


\begin{figure}[h]
    \centering
    \includegraphics[width=\linewidth]{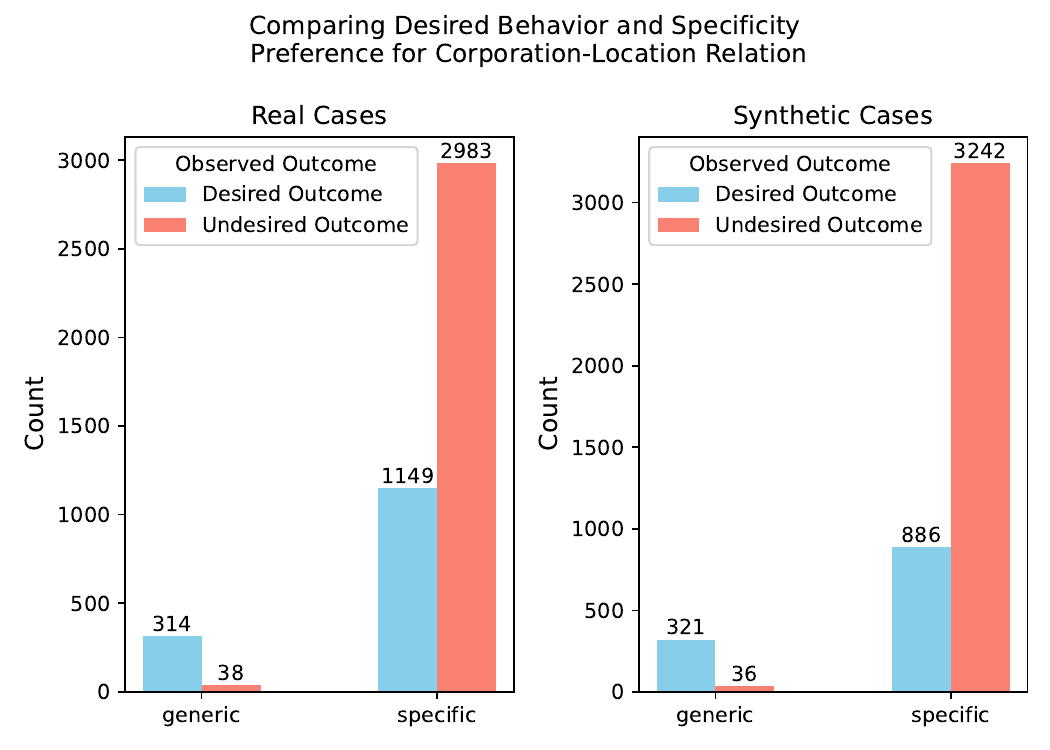}
    \caption{Counts of desired and undesired behaviors
across real and synthetic cases for Corporation-Location relation}
\end{figure}

\begin{figure}[h]
    \centering
    \includegraphics[width=\linewidth]{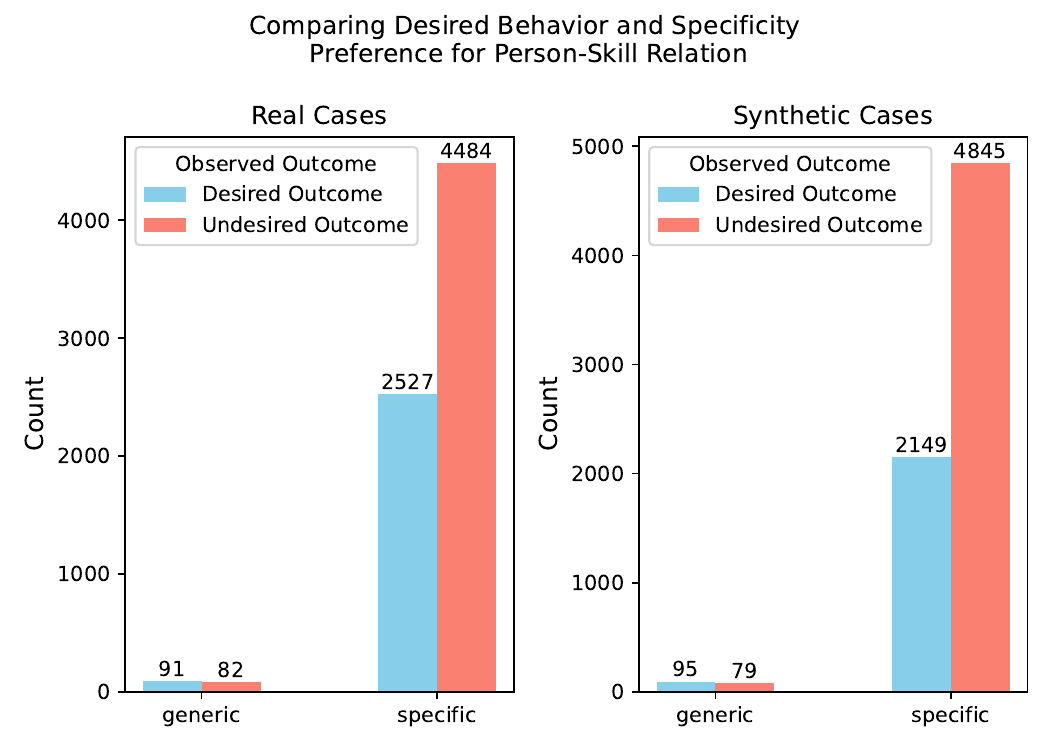}
    \caption{Counts of desired and undesired behaviors
across real and synthetic cases for Person-Skill relation.}
\end{figure}

\begin{figure}[h]
    \centering
    \includegraphics[width=\linewidth]{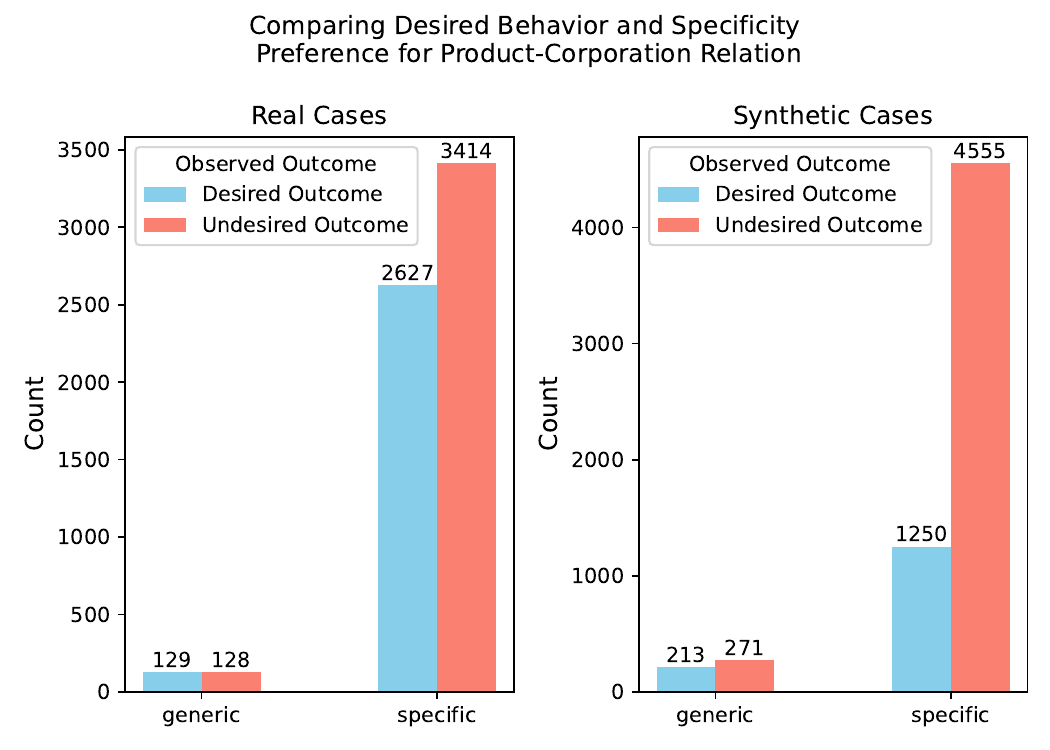}
    \caption{Counts of desired and undesired behaviors
across real and synthetic cases for Product-Corporation relation}
\end{figure}


\begin{figure}[h]
    \centering
    \includegraphics[width=\linewidth]{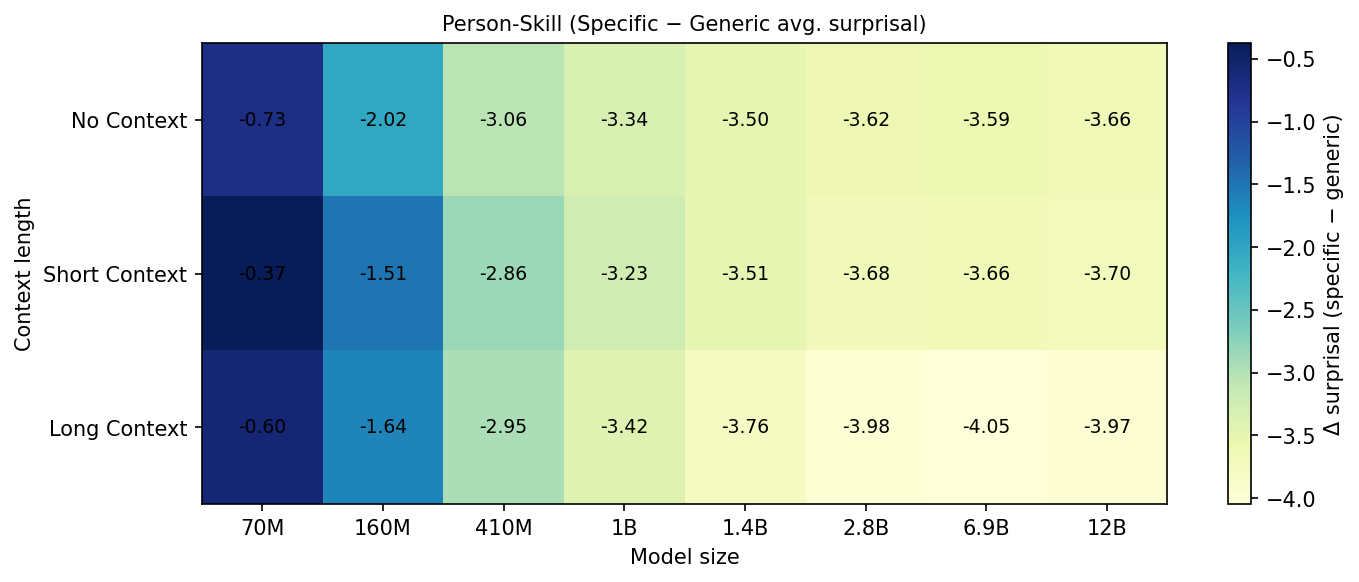}
    \caption{Comparison for Avg. Specific - General surprisal for different context length across models for Person-Skill relation.}
\end{figure}

\begin{figure}[h]
    \centering
    \includegraphics[width=\linewidth]{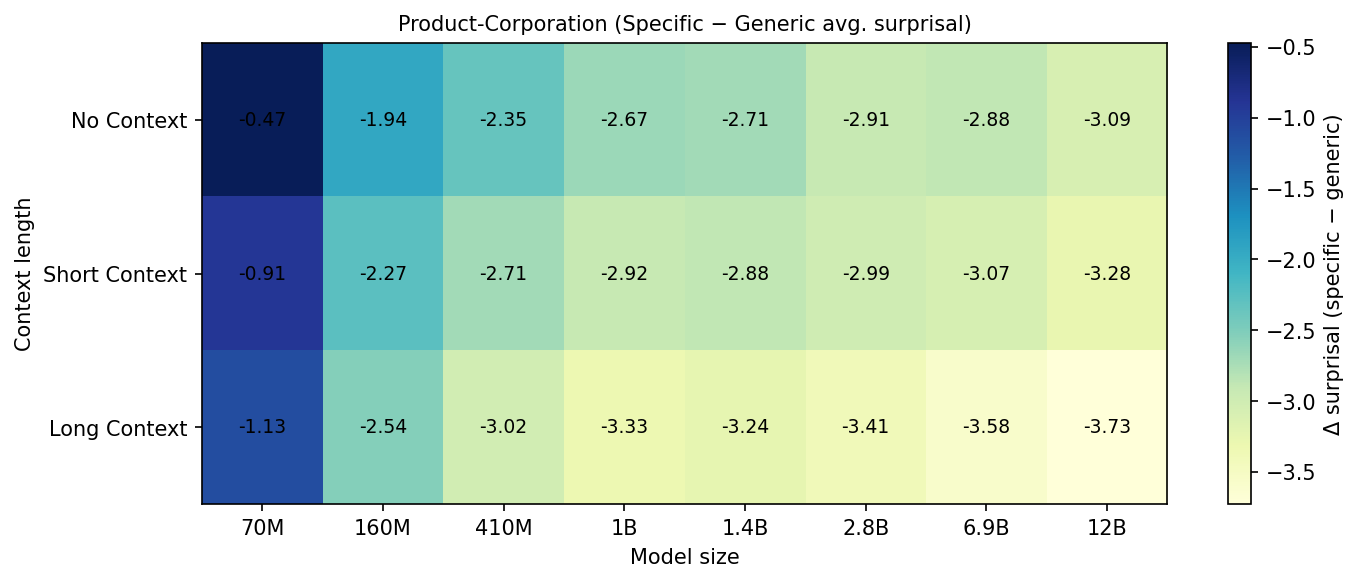}
    \caption{Comparison for Avg. Specific - General surprisal for different context length across models for  Product-Corporation relation.}
\end{figure}

\begin{figure}[h]
    \centering
    \includegraphics[width=\linewidth]{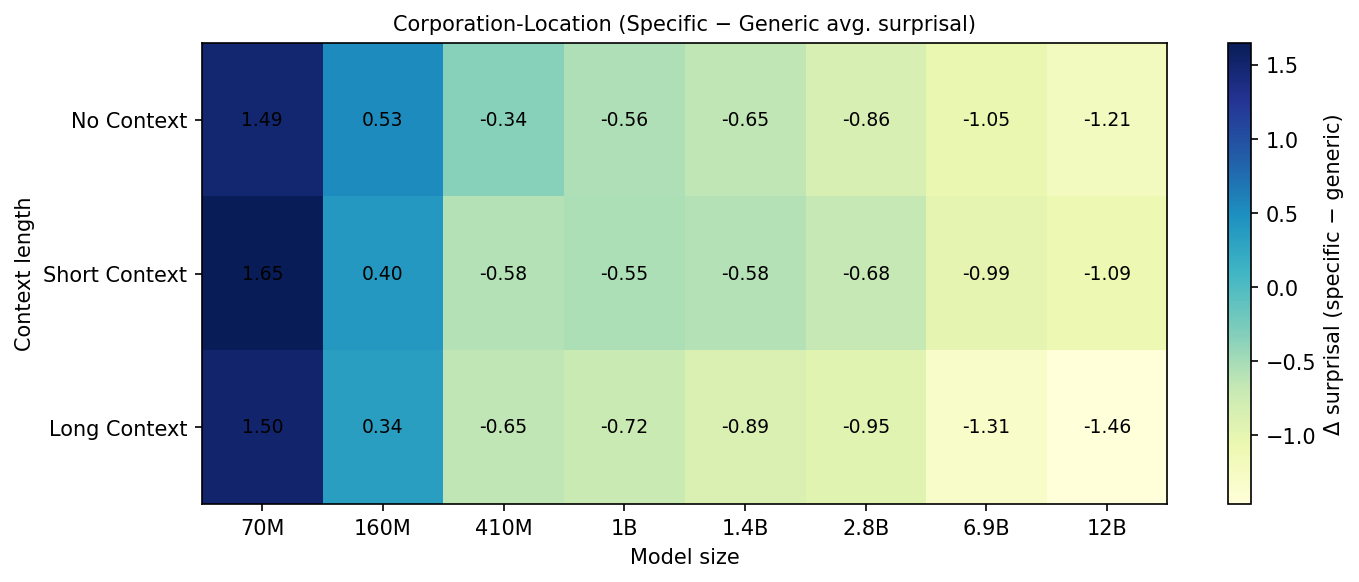}
    \caption{Comparison for Avg. Specific - General surprisal for different context length across models for Corporation-Location relation.}
\end{figure}

\begin{figure}[h]
    \centering
    \includegraphics[width=\linewidth]{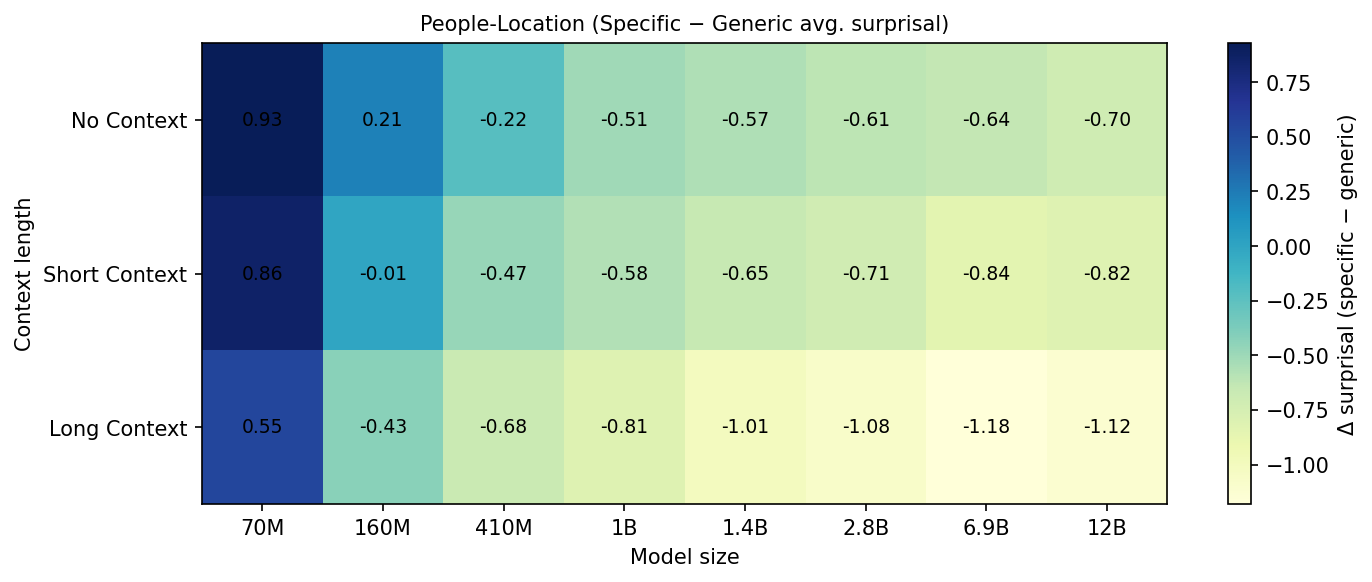}
    \caption{Comparison for Avg. Specific - General surprisal for different context length across models for People-Location relation.}
\end{figure}

\end{document}